\pdfoutput=1
\documentclass[10pt,twocolumn,letterpaper]{article}

\usepackage[pagenumbers]{cvpr}

\makeatletter
\iftoggle{cvprfinal}{}{%
  \IfFormatAtLeastTF{2025-06-01}{%
    \RemoveFromHook{build/column/before}[lineno]%
    \AddToHook{build/column/after}{%
      \setbox\@outputbox\vbox{%
        \boxmaxdepth\@maxdepth
        \protected@write\@auxout{}{\string\def\string\@LN@column{\if@firstcolumn1\else2\fi}}%
        \box\@outputbox}}%
  }{}%
}
\makeatother

\usepackage{cuted}

\usepackage[utf8]{inputenc}
\usepackage[T1]{fontenc}
\usepackage{mathtools}
\usepackage{microtype}
\usepackage{siunitx}
\usepackage{tabularx}
\usepackage{longtable}
\usepackage{placeins}
\usepackage{listings}
\usepackage{colortbl}  %
\definecolor{LightCyan}{HTML}{CCEEFF}
\usepackage{tcolorbox}
\tcbuselibrary{skins,breakable}
\definecolor{rootline}{HTML}{86AE5C}\definecolor{rootfill}{HTML}{D9E8C0}
\definecolor{rootback}{HTML}{F4F9EC}\definecolor{rootdark}{HTML}{2F5E1A}
\definecolor{subline}{HTML}{77AADD}\definecolor{subfill}{HTML}{D2E0F4}
\definecolor{subback}{HTML}{F1F6FC}\definecolor{subdark}{HTML}{1F4A80}
\definecolor{plainline}{HTML}{9C9C9C}\definecolor{plainfill}{HTML}{E4E4E4}
\definecolor{plainback}{HTML}{F6F6F6}\definecolor{plaindark}{HTML}{3A3A3A}
\definecolor{promptfield}{HTML}{AA3377}
\colorlet{promptdark}{black}
\newcommand{\promptfield}[1]{{\color{promptfield}\bfseries\{#1\}}}
\lstdefinestyle{prompt}{basicstyle=\scriptsize\ttfamily,columns=fullflexible,keepspaces=true,
  breaklines=true,breakatwhitespace=true,breakindent=0pt,upquote=true,frame=none,aboveskip=0pt,belowskip=0pt,
  moredelim=[l][\color{promptdark}]{\{"reason"},moredelim=[s][\color{promptdark}]{`}{`},
  literate={\{round\}}{{\promptfield{round}}}{7} {\{max\_rounds\}}{{\promptfield{max\_rounds}}}{12}
    {\{last\}}{{\promptfield{last}}}{6} {\{duration\}}{{\promptfield{duration}}}{10}
    {\{question\}}{{\promptfield{question}}}{10} {\{options\}}{{\promptfield{options}}}{9}
    {\{reports\}}{{\promptfield{reports}}}{9} {\{task\}}{{\promptfield{task}}}{6}
    {\{query\}}{{\promptfield{query}}}{7} {\{segments\}}{{\promptfield{segments}}}{10}
    {\{budget\}}{{\promptfield{budget}}}{8} {\{shown\}}{{\promptfield{shown}}}{7}
    {\{max\_read\}}{{\promptfield{max\_read}}}{10} {\{top\_k\}}{{\promptfield{top\_k}}}{7}
    {\{max\_evidence\}}{{\promptfield{max\_evidence}}}{14} {\{depth\}}{{\promptfield{depth}}}{7}
    {\{max\_depth\}}{{\promptfield{max\_depth}}}{11} {\{spawn\}}{{\promptfield{spawn}}}{7}
    {\{n\}}{{\promptfield{n}}}{3} {\{glance\}}{{\promptfield{glance}}}{8}
    {\{spacing\}}{{\promptfield{spacing}}}{9} {\{reply\}}{{\promptfield{reply}}}{7}
    {\{report\}}{{\promptfield{report}}}{8} {\{k\}}{{\promptfield{k}}}{3}
    {\{t0\}}{{\promptfield{t0}}}{4} {\{t1\}}{{\promptfield{t1}}}{4}}

\newenvironment{promptbox}[3]{%
  \colorlet{promptdark}{#1dark}%
  \begin{tcolorbox}[skin=bicolor,breakable,arc=3pt,boxrule=0.8pt,colframe=#1line,colback=#1back,colbacklower=white,
    colbacktitle=#1fill,coltitle=#1dark,fonttitle=\sffamily\bfseries\small,toptitle=1pt,bottomtitle=1pt,
    titlerule=0.8pt,segmentation style={draw=none},overlay={\ifnum\tcbsegmentstate=1\relax\draw[#1line,dashed,line width=0.6pt](segmentation.west)--(segmentation.east);\fi},left=6pt,right=6pt,top=5pt,bottom=5pt,
    middle=4pt,title={\raisebox{-0.42em}{\csname promptrobot#1\endcsname}\enspace #2\hfill{\boldmath#3}}]}
  {\end{tcolorbox}}
\newenvironment{promptmsg}{\par\noindent\begin{minipage}{\linewidth}}{\end{minipage}\par}
\newcommand{\promptpart}[2]{\noindent{\sffamily\scriptsize\bfseries\color{promptdark}#1}%
  {\sffamily\scriptsize\color{black!55}\enspace#2}\par\vspace{2pt}}
\newcommand{\prompttt}[1]{{\fontfamily{zi4}\selectfont #1}}
\newcommand{\promptnote}[1]{\par\vspace{2pt}\noindent{\sffamily\scriptsize\color{black!55}#1}}

\usepackage[ruled,vlined,linesnumbered]{algorithm2e}
\SetArgSty{textnormal}
\SetFuncSty{textsc}
\SetCommentSty{textnormal}
\SetKwComment{Comment}{$\triangleright$\ }{}

\definecolor{firebrick}{rgb}{0.698,0.133,0.133}

\newcommand{\limlabel}[1]{\hypertarget{lim:#1}{\textbf{\textcolor{firebrick}{(#1)}}}}
\newcommand{\limref}[1]{\hyperlink{lim:#1}{\textbf{\textcolor{firebrick}{(#1)}}}}

\newcommand{\method}{\textsc{VideoDispatch}\xspace}
\newcommand{\backbone}{GPT-6-Luna\xspace}

\definecolor{cvprblue}{rgb}{0.21,0.49,0.74}
\usepackage[pagebackref,breaklinks,colorlinks,allcolors=cvprblue]{hyperref}
\hypersetup{pdftitle={VideoDispatch: Global Planning and Local Searching for Long Video Understanding},
  pdfauthor={Chenwei Xu, Rahul Ghosh, Han Liu}}
\usepackage[capitalize]{cleveref}

\crefname{algocf}{Alg.}{Algs.}
\Crefname{algocf}{Algorithm}{Algorithms}

\def\confName{CVPR}
\def\confYear{2027}

\title{\method: Global Planning and Local Searching for Long Video Understanding}

\author{
Chenwei Xu\textsuperscript{1}\footnotemark\quad
Rahul Ghosh\textsuperscript{2}\quad
Han Liu\textsuperscript{1}\\[0.4em]
\textsuperscript{1}Northwestern University\quad
\textsuperscript{2}Amazon
}

\begin{document}
\maketitle

\begin{strip}
    \centering
    \vspace{\dimexpr-0.4in-12pt}
    \includegraphics[width=\linewidth]{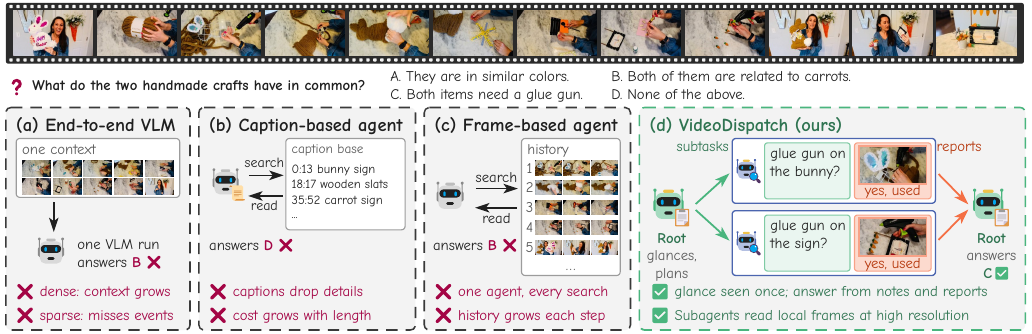}
    \captionsetup{hypcap=false}
    \captionof{figure}{\textbf{Comparison between \method and other methods.}
    (a) An end-to-end VLM reads uniformly sampled frames from the whole video.
    (b) An LLM agent searches a caption base.
    (c) A VLM agent searches and reads frames.
    (d) \method: the Root agent plans from a low-resolution glance and gives each subtask to a Subagent with a new context.
    The Root agent answers from notes and reports.}
    \captionsetup{hypcap=true}
    \label{fig:comparsion}
    \vspace{\dimexpr-0.05in-4pt}
\end{strip}
{\renewcommand\thefootnote{\fnsymbol{footnote}}\footnotetext[1]{Correspondence to \texttt{cxu@u.northwestern.edu}.}}

\begin{abstract}
We present \method, a multi-agent pipeline for agentic long video understanding. 
\method contains two parts: a Root agent and Subagents. 
The Root agent first takes a glance at the whole video by seeing low-resolution frames and writes notes about those seen frames. 
The Root agent also decides which frames to keep for supporting the final answer.  
It then dispatches several Subagents with different subtasks. 
Each subtask is a natural-language instruction that tells the Subagent what to find. 
The Subagent first uses CLIP embedding similarity to find the frames that best match its instruction. 
It views these frames and decides whether to search again, view more frames, or answer the subtask. 
Once it answers, it returns the answer to the Root agent together with the supporting frames.
Finally, the Root agent answers the question using its own notes and kept frames from the glance, together with the Subagents’ answers and supporting frames.
On three long-video benchmarks with the same backbone for all methods, \method matches the strongest video agents while using fewer tokens. 
At an equal token budget, it also outperforms uniform frame sampling on LVBench and LongVideoBench long.

\end{abstract}

\vspace{-40.8pt}
\section{Introduction}
\label{sec:intro}
\begin{figure*}[t]
    \centering
    \includegraphics[width=\textwidth]{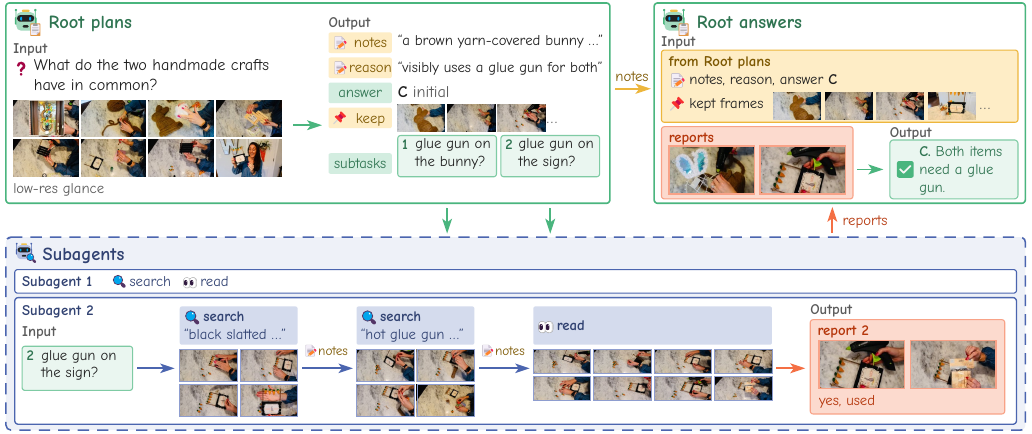}
    \vspace{-0.1in}
    \caption{\textbf{The \method Pipeline.}
    Each box shows the input and output of one run of the same VLM.
    In the planning stage (Root plans), the Root agent dispatches subtasks.
    Each Subagent solves one subtask in its own context, in parallel with the others.
    In the answering stage (Root answers), the Root agent sees only its notes, kept frames, and the reports.
    Frames and quotes come from a recorded \backbone run; here the initial answer is already correct.}
    \label{fig:overview}
    \vspace{-0.1in}
\end{figure*}

Agentic video understanding asks an agent to select the right frames for a question.
This is hard because videos are high-dimensional, temporally redundant, and semantically intricate~\cite{xu2026towards}.
End-to-end vision-language models (VLMs) read uniformly sampled frames of the whole video~\cite{bai2025qwen3,chen2025longvila,shu2025video,shen2024longvu} (\cref{fig:comparsion}a).
They face a trade-off: dense sampling needs a long context, and sparse sampling misses short events.
Agentic methods~\cite{fan2024videoagent,chen2025lvagent,liu2026videomind,tian2025ego,eltahir2026videoatlas,xu2026towards} instead steer large language models (LLMs) or VLMs to search long videos for short events.
They use multi-step reasoning and tools to decide where to look and when to answer.
Two paradigms exist.
The first converts video clips into captions and lets an LLM search and reason over the text~\cite{zhang2024simple,wang2024videoagent,wang2025videotree,ma2025drvideo,pang2026mr,zhang2026deep} (\cref{fig:comparsion}b).
Captions, however, may miss fine visual details.
The second lets a VLM agent search the video and read the frames it selects~\cite{wang2025active,yuan2025videoexplorer,li2026lenswalk} (\cref{fig:comparsion}c).
In most such methods, one agent runs all searches and keeps every observation in its context.
This causes two limitations.
\textbf{\limlabel{L1} Context Overload}: earlier observations pile up in the context and make every later run more expensive.
\textbf{\limlabel{L2} Global--Local Conflict}: locating a short event needs an overview of the whole video, while reading its details needs dense frames from a short span.
One context must carry both at every step.

Recursive Language Models (RLMs)~\cite{zhang2025recursive} keep a long context in an external environment and pass only short text between model runs.
Following this idea, we address \limref{L1} by replacing most seen frames with text in later runs.
We present \method (\cref{fig:comparsion}d), which has a Root agent and Subagents.
The Root agent glances at the video once, writes notes, and dispatches subtasks.
Each subtask is an instruction, and each Subagent solves one in a new context.
A Subagent finds candidate frames with CLIP~\cite{radford2021learning} or samples them uniformly.
It then searches again, reads frames around a candidate or across an interval, or answers.
It returns a report with timestamped evidence frames.
The Root agent answers from its notes, a few kept glance frames, and the reports, without seeing the whole glance again.

For \limref{L2}, \method splits the global and the local view by resolution.
The glance uses sparse low-resolution frames, which are cheap but show where the evidence may lie.
The Subagents then read frames at a higher resolution.

\method thus rests on one rule: the glance is seen once, and every later run receives text and a few chosen frames.
AVP~\cite{wang2025active} also moves from a coarse view to local observations, but its planner keeps the findings of every round in one context.
The Workers of VideoAtlas~\cite{eltahir2026videoatlas} zoom into a fixed frame grid level by level, in 153 to 174 model runs per question (\cref{tab:main}).
Our ablations show that the rule matters more than the split into agents.
Answering from the notes instead of the glance gives the whole saving, and the local search gives the accuracy.
A coarse plan suffices, and separate contexts save only a tenth of the input.

Our contribution is twofold.
(I) Methodologically, we propose \method, a training-free method for \limref{L1} and \limref{L2}.
A Root agent glances at the whole video once at low resolution, writes notes, and dispatches subtasks.
Subagents solve them in new contexts at a higher resolution and return reports with evidence frames.
The Root agent answers from its notes and the reports without seeing the glance again.
(II) Experimentally, we report video agents on the same backbone as \method on LVBench~\cite{wang2025lvbench}, Video-MME long~\cite{fu2025video}, and the long split of LongVideoBench~\cite{wu2024longvideobench}.
We count the model runs, tokens, and frames of every method.
\method matches the accuracy of the strongest agents with 27 to 34\% fewer tokens than AVP.
At equal input tokens, it beats uniform sampling on LVBench and LongVideoBench long. 
We summarize our methods and provide a qualitative example in \cref{fig:overview} and \cref{fig:example}.

\section{Related Work}
\label{sec:related}
\noindent \textbf{Long Video Understanding.}
Long-video benchmarks~\cite{fu2025video,wu2024longvideobench,zhou2025mlvu,wang2025lvbench} extend video question answering from short clips to videos of minutes to hours.
End-to-end VLMs handle the many visual tokens by extending the context~\cite{chen2025longvila,bai2025qwen3,comanici2025gemini}, compressing tokens~\cite{shu2025video,shen2024longvu}, or selecting keyframes~\cite{sun2025mdp3,ye2025re}.
All of them still read the sampled video in one context.
\method instead gives a frozen VLM a sparse glance and then only the frames each subtask selects.

\noindent \textbf{Agentic Frameworks for Long Video Understanding.}
Many agentic frameworks~\cite{zhang2024simple,wang2024videoagent,wang2025videotree,zhang2025silvr,wang2026thinkverify} split perception from reasoning.
A captioner converts the video into text, and an LLM searches the captions to answer.
Other works~\cite{fan2024videoagent,ma2025drvideo,pang2026mr,zuo2026videolucy,yan2026ava,choi2026keep,xu2026finding} combine captions or a text memory with a VLM that inspects selected frames.
Tool-based agents~\cite{zhang2026deep,yuan2025videodeepresearch,lin2026videoseek} give a reasoning model tools to search the video, and several works~\cite{yuan2025videoexplorer,tian2025ego,liu2026videomind} train planners to use grounding tools.
Since captions miss fine details, recent studies~\cite{yang2025vca,wang2025active,li2026lenswalk,huang2026temporal,wang2026videochat} let a VLM search the frames directly.
Multi-agent frameworks~\cite{chen2025lvagent,xu2026multi} let several VLM agents collaborate.
AVP~\cite{wang2025active} plans, observes, and reflects for each question and keeps its findings as timestamped text across rounds.
LensWalk~\cite{li2026lenswalk} keeps a text summary of each observation.
In both, one context holds the history of all searches.
In \method, each Subagent holds only the history of its own subtask and returns its answer with the source frames.
Because its Root agent answers from notes instead of the glance, \method uses fewer tokens than AVP at similar or higher accuracy (\cref{sec:exp_ablation}).

\noindent \textbf{Recursive Language Models and Context Management.}
The context of an agent grows with every step.
ReAct~\cite{yao2022react} reasons and acts in one context, and multi-turn visual agents append the images of every turn.
ReViSe~\cite{xu2026towards} bounds this growth by passing only a written summary between rounds.
MEM1~\cite{zhou2026mem1} trains an agent to merge its memory and each observation into one compact state.
RLMs~\cite{zhang2025recursive} instead keep a long prompt in an external environment and delegate parts of it to further runs of the model.
VideoAtlas~\cite{eltahir2026videoatlas} applies RLMs to video: a Master assigns cells of a hierarchical frame grid to Workers, which expand them level by level.
Chain-of-Agents~\cite{zhang2024chain} passes chunks of a long input through a chain of workers, and ReWOO~\cite{xu2023rewoo} plans tool calls before running them.
\method differs in two ways.
First, its input is bounded: a question sees at most 328 frames for any video length, and notes replace the frames already read.
Second, its search adapts: each Subagent searches with a CLIP index and reads the spans it picks at a rate it chooses.

\section{Method}
\label{sec:method}
For each question, \method takes a global view of the video, runs local searches with Subagents, and answers from both (\cref{fig:overview}).
The Root agent looks at a low-resolution glance of the whole video, writes notes on it, and dispatches subtasks (\cref{sec:root_plan}).
Each Subagent searches and reads frames for its subtask and returns a report with evidence frames (\cref{sec:sub}).
The Root agent then answers from its notes, a few kept glance frames, and the reports, without seeing the whole glance again.
A frame budget caps the distinct frames of a question, and reason-as-memory stops each turn from resending earlier frames (\cref{sec:budget}).
\Cref{alg:root,alg:sub} summarize the two procedures with their limits.

\begin{figure*}[t]
\begin{minipage}[t]{0.4375\textwidth}
\csname @twocolumnfalse\endcsname
\begin{algorithm}[H]
\small
\caption{\method: Root agent}
\label{alg:root}
\DontPrintSemicolon
\SetKwFunction{Sub}{Sub}
\KwIn{video $V$ of duration $T$,\newline question $q$, options $\mathcal{Y}$}
\KwOut{answer $\hat{y}^\mathrm{final}\in\mathcal{Y}$}
$\Gamma\gets$ $G$ uniform low-resolution frames of $V$ \Comment*[r]{glance}
$(m,\hat{y}^\mathrm{init},\mathcal{K},\mathcal{Q})\gets\pi_\theta^{\mathrm{root}}(\Gamma,q,\mathcal{Y})$ \Comment*[r]{planning}
$\mathcal{Q}\gets$ the first $M$ valid subtasks of $\mathcal{Q}$\;
\lIf{$\mathcal{Q}=\varnothing$}{\Return $\hat{y}^\mathrm{init}$}
\ForEach{$u_k\in\mathcal{Q}$, in parallel}{
  $r_k\gets$ \Sub{$u_k,q,\mathcal{Y}$}\;
}
\Return $\pi_\theta^{\mathrm{root}}\bigl(q,\mathcal{Y},m,\hat{y}^\mathrm{init},$\newline $\{x_t\}_{t\in\mathcal{K}},\{(u_k,r_k)\}_k\bigr)$\;
\vspace{3.26pt}%
\end{algorithm}
\end{minipage}\hfill
\begin{minipage}[t]{0.53\textwidth}
\csname @twocolumnfalse\endcsname
\begin{algorithm}[H]
\small
\caption{\method: Subagent}
\label{alg:sub}
\DontPrintSemicolon
\SetKwFunction{Sub}{Sub}
\SetKwProg{Fn}{procedure}{}{}
\Fn{\Sub{$u,\,q,\,\mathcal{Y}$}}{
  $b\gets B$; $o_1\gets(q,\mathcal{Y},u)$ \Comment*[r]{budget}
  \For{$\ell=1,\dots,R$}{
    $z_\ell\gets\pi_\theta^{\mathrm{sub}}\bigl(\{(\tilde{o}_i,z_i)\}_{i<\ell},\{x_t\}_{t\in\mathcal{K}_{\ell-1}},o_\ell\bigr)$\;
    \lIf{$a_\ell=\textsc{Return}$ \textbf{or} $\ell=R$ \textbf{or} $b\le 0$}{\Return report $r$}
    \uIf{$a_\ell=\textsc{Read}(w,f)$}{
      $o_{\ell+1}\gets$ at most $\min(n_r,b)$ uniform frames of $w$\;
    }
    \uElseIf{$a_\ell=\textsc{Search}(c,w)$}{
      $o_{\ell+1}\gets$ the $\min(k,b)$ strongest new peaks in $w$\;
    }
    \lElse{$o_{\ell+1}\gets$ an error message}
    $b\gets b-|o_{\ell+1}|$ \Comment*[r]{frames shown}
  }
}
\end{algorithm}
\end{minipage}
\end{figure*}

\subsection{Root Agent}
\label{sec:root_plan}
The inputs are a video $V$ of duration $T$ seconds, a question $q$, and a set $\mathcal{Y}$ of answer options.
We write $x_t$ for the frame at time $t\in[0,T]$.
The Root agent first sees a glance $\Gamma$ of $G$ uniformly sampled frames of the whole video.
It plans subtasks from the glance and dispatches them to Subagents.

\noindent \textbf{Planning.}
The Root agent receives the question $q$, the options $\mathcal{Y}$, and the glance $\Gamma$.
It returns notes on the video, an initial answer, a few glance frames to keep, and subtasks for the Subagents.
We write $\pi_\theta$ for the VLM with parameters $\theta$, and $\pi_\theta^{\mathrm{root}}$ and $\pi_\theta^{\mathrm{sub}}$ for the Root agent and the Subagent.
The notes $m$ describe what happens in the video and when.
For example, ``two men in a dim basement (0--78\,s), then Peter and his mother at her apartment (80--195\,s).''
The initial answer $\hat{y}^\mathrm{init}\in\mathcal{Y}$ is the option that the glance supports best.
The keep list $\mathcal{K}$ holds up to $n_g$ glance frames that the Root agent sees again when it answers.
$\mathcal{Q}$ is the set of at most $M$ subtasks.
This first turn of the Root agent, the planning stage, is thus
\begin{equation}
(m,\,\hat{y}^\mathrm{init},\,\mathcal{K},\,\mathcal{Q}) \;=\; \pi_\theta^{\mathrm{root}}\bigl(\Gamma,\,q,\,\mathcal{Y}\bigr).
\label{eq:call1}
\end{equation}
Without a valid subtask, the Root answers with $\hat{y}^\mathrm{init}$.

\noindent \textbf{Subtasks.}
A subtask $u$ is an instruction for one Subagent.
It tells the Subagent what to look for and what to report, for example, ``Count the times a person opens a door; give the time of each.''
The Subagent does not see the glance, so the instruction describes what the target looks like.
The Subagent then chooses which frames to read and how densely.

\noindent \textbf{Answering.}
\label{sec:root_answer}
Once the Subagents return their reports, the Root agent answers the question in the answering stage.
It receives $q$, $\mathcal{Y}$, the notes $m$, the initial answer $\hat{y}^\mathrm{init}$, and the kept glance frames $\{x_t:t\in\mathcal{K}\}$.
It also receives each subtask $u_k\in\mathcal{Q}$ with its report $r_k$ (\cref{sec:sub}).
The Root agent $\pi_\theta^{\mathrm{root}}$ then gives the final answer
\begin{equation}
\hat{y}^\mathrm{final} \;=\; \pi_\theta^{\mathrm{root}}\Bigl(q,\,\mathcal{Y},\,m,\,\hat{y}^\mathrm{init},\,\{x_t : t\in\mathcal{K}\},\,\{(u_k,r_k)\}_{k=1}^{|\mathcal{Q}|}\Bigr).
\label{eq:call2}
\end{equation}
The answering stage thus sees the notes and a few kept glance frames instead of the whole glance.
We call this answering from notes. It applies reason-as-memory (\cref{sec:memory}) to the two turns of the Root agent.
The Root agent uses the reports to verify or revise its initial answer.
If the reports give no evidence against it, the Root agent keeps the initial answer.

\noindent \textbf{Example.}
In \cref{fig:overview}, the question asks what two handmade crafts have in common.
The glance shows a yarn-covered bunny and a sign, and the Root agent gives C as its initial answer.
It asks two Subagents whether a glue gun is used on each craft.
Subagent 2 searches twice, reads the frames around a candidate, and reports that a glue gun is used on the sign.
With both reports, the answering stage keeps C.

\subsection{Subagent}
\label{sec:sub}
Each subtask goes to its own Subagent.
The Subagent receives the question, the options, and the instruction $u$ as context.
Like a sub-call of an RLM, it starts in a new context, and all Subagents run in parallel.
In each turn, it writes a reason and takes one action.
\Cref{sec:memory} describes what a turn keeps from the earlier turns of its Subagent.

\noindent \textbf{Video Environment.}
Before any question arrives, CLIP~\cite{radford2021learning} embeds the frames of each video at \SI{1}{fps}.
The Subagent picks the window $w$ of each search or read.
\textsc{Search}$(c,w)$ scores the indexed frames in $w$ by the cosine similarity of their image embeddings to the text embedding of the caption $c$.
It shows the $k$ strongest peaks, at least $\delta$ seconds apart, in time order.
It leaves out every moment within $\delta/2$ seconds of a frame that earlier searches showed.
\textsc{Read}$(w,f)$ shows frames with uniform spacing in $w$ at frame rate $f$, at most $n_r$ per read.
\textsc{Return}$(y,\mathcal{H},E,U)$ ends the Subagent, where $E$ holds the indices of at most $n_e$ evidence frames.
Each local frame is a 512-pixel image labeled with its index, time, and source, and costs one unit of the budget $B$.
A Subagent returns when its budget or its $R$ replies run out.

\noindent \textbf{Reports and Evidence Frames.}
A Subagent returns a report $r=(y,\mathcal{H},\mathcal{E},U,\sigma)$.
It holds a free-text answer $y$, the times $\mathcal{H}$ that the answer refers to, and at most $n_e$ timestamped evidence frames $\mathcal{E}$.
It also lists the points $U$ that its frames do not settle and a flag $\sigma$ that tells whether its limits stopped it.
Code decodes the cited frames again at their timestamps and attaches them to the report.
The answering stage shows each report as text, followed by its evidence frames.
The Root agent can thus check each report against its frames and merge the findings of all subtasks into one timeline.

\subsection{Frame Budget and Reason-as-Memory}
\label{sec:budget}
Two limits keep a question cheap.
The frame budget caps the distinct frames per question.
Reason-as-memory and answering from notes cap how often they are shown again.

\noindent \textbf{Frame Budget.}
Let $S$ be the set of distinct frames that the VLM sees for one question.
The glance holds $G$ frames, the Root agent dispatches at most $M$ subtasks, and each Subagent decodes at most $B$ local frames.
Kept glance frames and evidence frames were already decoded, so $|S| \le G + M\,B$.
With $G=256$, $M=3$, and $B=24$, a question sees at most 328 frames for a video of any length: 256 at low resolution and 72 at 512 pixels.
Only the CLIP index grows with the video, and all questions about the video share it.

\noindent \textbf{Reason-as-Memory.}
\label{sec:memory}
A standard multi-turn chat resends every earlier image in every turn.
Reason-as-memory keeps earlier observations as text instead, which bounds the images of a turn.
Each reply $z_\ell=(m_\ell,a_\ell,\mathcal{K}_\ell)$ of a Subagent holds a reason $m_\ell$, an action $a_\ell$, and an optional keep list $\mathcal{K}_\ell$.
The reason describes the new frames by index, and the keep list names at most $n_k$ earlier frames to see again.
Let $o_\ell$ be the observation of turn $\ell$ and $\tilde{o}_\ell$ its text version, which keeps the frame labels and drops the images.
The model turn at turn $\ell$ returns
\begin{equation}
z_\ell \;=\; \pi_\theta^{\mathrm{sub}}\Bigl(\{(\tilde{o}_i,z_i)\}_{i<\ell},\ \{x_t : t\in\mathcal{K}_{\ell-1}\},\ o_\ell\Bigr).
\end{equation}
A turn shows only the kept frames and the new observation, at most $n_k+\max(n_r,k)=12$ images by default.

\noindent \textbf{Answering from Notes.}
The Root agent applies the same idea across its two turns.
The notes $m$ replace the glance frames, and the keep list $\mathcal{K}$ pins the ones that matter most.
The answering stage thus takes 3K to 4K input tokens instead of the 31K of a turn with the whole glance (\cref{sec:exp_ablation}).

\begin{figure*}[t]
\centering
\includegraphics[width=\linewidth]{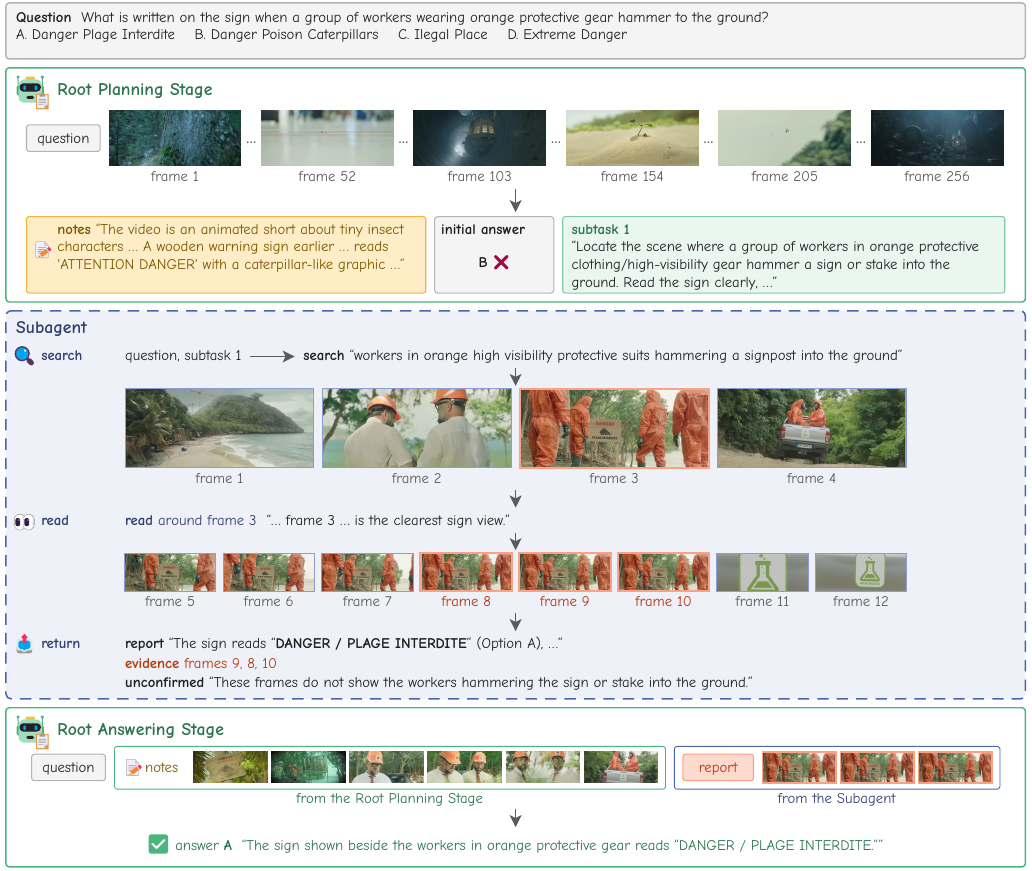}
\vspace{-0.05in}
\caption{\textbf{Qualitative Example.}
In the planning stage, the Root agent sees the question and the glance.
It finds the workers in orange gear but cannot read the sign.
Its notes describe another sign with a caterpillar graphic, and its initial answer.
It dispatches one subtask: find the workers and read the sign.
The Subagent receives the question and this subtask. 
It then decides to search frames that best match the queries. 
This search result shows the sign.
It then reads more frames around that result and returns a report.
The report summarizes the findings, using frames 9, 8, and 10 as evidence, and states what is not confirmed.
In the answering stage, the Root agent sees the notes, kept glance frames, and the report with its evidence frames.
It changes the answer based on notes from the Root agent and reports from Subagents.}
\label{fig:example}
\end{figure*}

\section{Experiments}
\label{sec:exp}
\begin{table*}[t]
\centering
\caption{\textbf{Comparison with Video Agents}.
Every method runs on \backbone, also for its captioning model. 
Acc: Accuracy in \%.  F: frame-based methods. C: Caption-based methods. F+C: Frame- and caption-based method. 
Runs: number of model calls.
Tokens: sum of input and output tokens.
Frames: number of distinct frames used, with captioning divided among the questions of a video. 
\method is the most accurate on LongVideoBench long, 5.0 points above AVP and 6.4 above VideoLucy.
On LVBench and Video-MME long, it is 0.5 points below VideoLucy and 0.4 below AVP, the best methods there.
Among frame-based methods, it is the most accurate on LVBench and LongVideoBench long.
It uses 27 to 34\% fewer tokens and 33 to 43\% fewer frames than AVP.
It uses less than 1/10 of the tokens of VideoLucy and less than 1/16 of those of VideoAtlas.
The other agents are 4.0 to 27.7 points below \method on every benchmark.}
\vspace{-0.08in}
\label{tab:main}
\setlength{\tabcolsep}{3.4pt}
\resizebox{0.85\textwidth}{!}{%
\begin{tabular}{lccccccccccccc}
\toprule
& & \multicolumn{4}{c}{LVBench} & \multicolumn{4}{c}{Video-MME long} & \multicolumn{4}{c}{LongVideoBench long} \\
\cmidrule(lr){3-6}\cmidrule(lr){7-10}\cmidrule(lr){11-14}
& & & \multicolumn{3}{c}{Cost $\downarrow$} & & \multicolumn{3}{c}{Cost $\downarrow$} & & \multicolumn{3}{c}{Cost $\downarrow$} \\
\cmidrule(lr){4-6}\cmidrule(lr){8-10}\cmidrule(lr){12-14}
Method & Input & Acc.\,$\uparrow$ & Runs & Tokens & Frames & Acc.\,$\uparrow$ & Runs & Tokens & Frames & Acc.\,$\uparrow$ & Runs & Tokens & Frames \\
\midrule
LLoVi~\cite{zhang2024simple} & C & 48.2 & 8.7 & 41.4K & 53.7 & 60.3 & 22.7 & 73.1K & 164.6 & 52.5 & 14.1 & 45.1K & 95.8 \\
VideoAgent~\cite{wang2024videoagent} & C & 38.0 & 9.4 & 6.8K & 8.8 & 56.3 & 9.0 & 7.2K & 9.7 & 47.7 & 9.7 & 7.8K & 9.2 \\
VideoLucy~\cite{zuo2026videolucy} & F+C & \textbf{60.2} & 37.5 & 525.2K & 761.5 & 61.9 & 41.4 & 564.2K & 1{,}260.3 & 62.6 & 46.0 & 591.1K & 978.9 \\
\midrule
AVP~\cite{wang2025active} & F & 58.6 & 4.0 & 65.6K & 416.6 & \textbf{64.7} & 4.0 & 76.3K & 501.1 & 64.0 & 4.1 & 72.1K & 406.6 \\
ReViSe~\cite{xu2026towards} & F & 32.0 & 5.0 & 9.7K & 10.1 & 47.0 & 5.0 & 9.8K & 10.7 & 43.3 & 5.1 & 10.1K & 10.8 \\
VideoAtlas~\cite{eltahir2026videoatlas} & F & 46.3 & 152.7 & 793.4K & 4{,}542.3 & 54.6 & 173.5 & 900.3K & 4{,}978.8 & 55.7 & 166.9 & 865.7K & 4{,}171.2 \\
LensWalk~\cite{li2026lenswalk} & F & 51.5 & 25.4 & 104.7K & 161.2 & 55.7 & 22.2 & 99.5K & 169.7 & 57.3 & 27.2 & 116.6K & 189.7 \\
\method & F & \cellcolor{LightCyan}59.7 & \cellcolor{LightCyan}6.7 & \cellcolor{LightCyan}47.8K & \cellcolor{LightCyan}274.6 & \cellcolor{LightCyan}64.3 & \cellcolor{LightCyan}8.5 & \cellcolor{LightCyan}54.4K & \cellcolor{LightCyan}285.6 & \cellcolor{LightCyan}\textbf{69.0} & \cellcolor{LightCyan}7.0 & \cellcolor{LightCyan}47.7K & \cellcolor{LightCyan}270.5 \\
\bottomrule
\end{tabular}}
\vspace{-0.12in}
\end{table*}

We compare \method with video agents and uniform sampling on three long-video benchmarks, all on the same backbone.
Ablations then show where its accuracy and its savings come from and what each of its parts costs.
We also show a qualitative example in \cref{fig:example}.

\subsection{Long-Video Question Answering}
\label{sec:exp_setup}
\noindent \textbf{Benchmarks and Metrics.}
We evaluate on three long-video benchmarks with multiple-choice questions.
LVBench~\cite{wang2025lvbench} has 1{,}549 questions on 103 videos, which last 68 minutes on average.
The long split of Video-MME~\cite{fu2025video} has 900 questions on 300 videos of 30 to 60 minutes.
The long split of the LongVideoBench~\cite{wu2024longvideobench} validation set (LongVideoBench long) has 564 questions on 188 videos of 15 to 60 minutes.
We do not use subtitles or audio.
We report accuracy and three costs per question: model runs (how many times a VLM is used), tokens (input and output), and distinct frames used.
We count any failed run as a wrong answer.

\noindent \textbf{Settings.}
We use \backbone for each method.
We use CLIP ViT-B/32~\cite{radford2021learning} embeds each video at \SI{1}{fps}.
In the planning stage, the Root agent sees a glance of $G=256$ frames and dispatches at most $M=3$ subtasks.
Each Subagent searches and reads with a budget of $B=24$ frames at 512 pixels.
A Subagent has at most $R=8$ replies and reads at most $n_r=8$ frames at a time.
A search shows $k=4$ peaks at least $\delta=\SI{10}{s}$ apart.
A Subagent keeps at most $n_k=4$ earlier frames and cites at most $n_e=3$ evidence frames, and the Root agent keeps at most $n_g=16$ glance frames.

\noindent \textbf{Baselines.}
We compare with seven training-free video agents with public code.
All run on \backbone with their own prompts and default settings, adapted where needed, for example with frames as images instead of a video input.
Four of them give the model only frames, like \method (input F in \cref{tab:main}).
AVP~\cite{wang2025active} plans, observes, and reflects on the video for each question.
It is the strongest of these agents on LVBench and Video-MME.
It keeps the frame caps of its code, at most 512 low-resolution frames per observation.
LensWalk~\cite{li2026lenswalk} lets a reasoning model use scan and inspection tools over several turns.
VideoAtlas~\cite{eltahir2026videoatlas} applies RLMs to video: a Master assigns cells of a frame grid to Workers, which zoom into them.
ReViSe~\cite{xu2026towards} shows at most three frames per round.
VideoLucy~\cite{zuo2026videolucy}, LLoVi~\cite{zhang2024simple}, and VideoAgent~\cite{wang2024videoagent} reason over captions from \backbone.
VideoLucy refines a memory for the question, LLoVi summarizes the whole video, and VideoAgent captions frames that CLIP retrieves.
We also compare with uniform sampling and a blind baseline.
Uniform-$N$ answers in one run from the middle frames of $N$ equal parts of the video, each the size of a glance frame and labeled with its time.
On LVBench, Uniform-256, Uniform-400, and Uniform-650 match the input tokens of \method with a glance of $G=128$, 256, and 512 frames.
Blind answers in one run from the question and the options alone.
It scores 23.9 on LVBench but 48.4 and 45.6 on the other two benchmarks.

\begin{figure}[t]
    \centering
    \includegraphics[width=\linewidth]{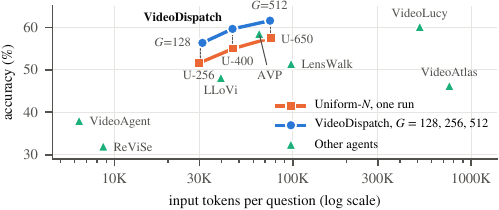}
    \vspace{-0.25in}
    \caption{\textbf{Accuracy against input tokens on LVBench.}
    All 1{,}549 questions; \method varies its glance of $G$ frames.
    Dotted lines join runs with about the same input tokens, where \method is significantly higher, by 4.8, 4.7, and 4.1 points.}
    \label{fig:pareto}
    \vspace{-0.15in}
\end{figure}

\begin{table}[t]
    \centering
    \caption{\textbf{Comparison with Uniform Sampling on LVBench.}
    Rows are grouped by similar tokens.
    $G$: glance frames of \method, 256 by default.
    Acc: accuracy in \%.
    Input: input tokens per question.
    At similar input tokens, \method is significantly more accurate than uniform sampling at every glance size.}
    \label{tab:uniform}
    \vspace{-0.05in}
    \footnotesize
    \setlength{\tabcolsep}{3pt}
    \begin{tabular*}{\linewidth}{@{\extracolsep{\fill}}lcc}
    \toprule
    Method & Acc.\,$\uparrow$ & Input $\downarrow$ \\
    \midrule
    Uniform-256 & 51.6 & 29.7K \\
    \method, $G=128$ & 56.4 & 31.2K \\
    \midrule
    Uniform-400 & 55.0 & 46.4K \\
    \method & 59.7 & 46.0K \\
    \midrule
    Uniform-650 & 57.5 & 75.2K \\
    \method, $G=512$ & 61.6 & 74.6K \\
    \bottomrule
    \end{tabular*}
    \vspace{-0.1in}
\end{table}

\noindent \textbf{Results.}
\method is both efficient and accurate (\cref{tab:main}).
It uses 27 to 34\% fewer tokens than AVP, all of the saving in input tokens.
VideoLucy and VideoAtlas use 10 to 12$\times$ and 17 to 18$\times$ its tokens.
\method reaches 59.7, 64.3, and 69.0 on LVBench, Video-MME long, and LongVideoBench long.
AVP, the strongest agent that reads only frames, reaches 58.6, 64.7, and 64.0.
The two are level on LVBench and Video-MME long, and \method is significantly higher on LongVideoBench long.
VideoLucy is level with \method on LVBench and lower on the other two benchmarks.
LensWalk, VideoAtlas, LLoVi, VideoAgent, and ReViSe stay significantly below \method everywhere, by 4.0 to 27.7 points.

\noindent \textbf{Uniform Sampling.}
With similar input tokens, \method is significantly more accurate than uniform sampling on LVBench (\cref{tab:uniform,fig:pareto}).
With a glance of $G=128$, 256, and 512 frames, it is 4.8, 4.7, and 4.1 points above Uniform-256, Uniform-400, and Uniform-650, respectively.

\noindent \textbf{Backbones Analysis.} 
\Cref{tab:qwen} compares \method with AVP on three backbones: the proprietary \backbone and GPT-6-Sol, and the open-source Qwen3.8-27B.
In both model families, \method is significantly more accurate than AVP on LVBench, with Qwen3.8-27B and GPT-6-Sol.
It uses 7 to 38\% fewer tokens than AVP on all three models.
On Video-MME long, the two achieve similar accuracy with every backbone.
With GPT-6-Sol, \method is also significantly above AVP on LongVideoBench long, 75.2 against 71.1.
A stronger backbone boosts the accuracy of \method more than it helps AVP.
From \backbone to GPT-6-Sol, \method improves by 11.8 points on LVBench and 11.7 on Video-MME long, and AVP by only 7.9 and 10.8.
The lead of \method over AVP on LVBench grows from 1.1 points with \backbone to 2.8 with Qwen3.8-27B and to 5.0 points with GPT-6-Sol.
The Subagents help with every backbone.
Over the initial answer, they add 7.3 to 10.6 points on LVBench and 2.5 to 6.6 on LongVideoBench long.
On Video-MME long, they change it by at most 1.8 points, as the glance already suffices.

\begin{table}[t]
    \centering
    \caption{\textbf{Comparison with AVP on Three Backbones.}
    We compare AVP under the proprietary model \backbone and GPT-6-Sol, and the open-source model Qwen3.8-27B.
    The rows of \backbone are same as \cref{tab:main}.
    LVB long: LongVideoBench long.
    Acc: accuracy in \%.
    Tokens: sum of input and output tokens.
    $^*$Initial answer, with the tokens of the planning stage.
    With Qwen3.8-27B and GPT-6-Sol, \method is significantly more accurate than AVP on LVBench.
    On Video-MME long, the two are level with every backbone.
    \method uses 7 to 38\% fewer tokens than AVP with every backbone.
    The Subagents add 7.3 to 10.6 points on LVBench with every backbone.}
    \label{tab:qwen}
    \vspace{-0.05in}
    \footnotesize
    \setlength{\tabcolsep}{1.8pt}
    \begin{tabular*}{\linewidth}{@{\extracolsep{\fill}}lcccccc}
    \toprule
    & \multicolumn{2}{c}{LVBench} & \multicolumn{2}{c}{Video-MME long} & \multicolumn{2}{c}{LVB long} \\
    \cmidrule(lr){2-3} \cmidrule(lr){4-5} \cmidrule(lr){6-7}
    Method & Acc.\,$\uparrow$ & Tokens\,$\downarrow$ & Acc.\,$\uparrow$ & Tokens\,$\downarrow$ & Acc.\,$\uparrow$ & Tokens\,$\downarrow$ \\
    \midrule
    \multicolumn{7}{l}{\emph{Qwen3.8-27B}} \\
    AVP~\cite{wang2025active} & 60.7 & 48.8K & 67.1 & 50.9K & 65.3 & 51.2K \\
    \method{}$^*$ & 53.2 & 22.6K & 64.2 & 22.5K & 61.5 & 22.4K \\
    \method & 63.6 & 42.1K & 65.1 & 47.5K & 68.1 & 41.4K \\
    \midrule
    \multicolumn{7}{l}{\emph{\backbone}} \\
    AVP~\cite{wang2025active} & 58.6 & 65.6K & 64.7 & 76.3K & 64.0 & 72.1K \\
    \method{}$^*$ & 52.4 & 31.5K & 64.6 & 31.3K & 66.5 & 30.7K \\
    \method & 59.7 & 47.8K & 64.3 & 54.4K & 69.0 & 47.7K \\
    \midrule
    \multicolumn{7}{l}{\emph{GPT-6-Sol}} \\
    AVP~\cite{wang2025active} & 66.5 & 69.9K & 75.4 & 80.0K & 71.1 & 80.2K \\
    \method{}$^*$ & 60.9 & 30.7K & 74.2 & 30.6K & 70.4 & 30.4K \\
    \method & 71.5 & 53.1K & 76.0 & 60.8K & 75.2 & 50.1K \\
    \bottomrule
    \end{tabular*}
    \vspace{-0.1in}
\end{table}

\subsection{Ablation Studies}
\label{sec:exp_ablation}
\begin{table}[t]
    \centering
    \caption{\textbf{Ablations.}
    Acc: accuracy in \%.
    Input: input tokens per question.
    Removing the Subagents or their reports costs 4.7 to 7.3 points, and every other change stays within 1.3 points.
    With an early answer, the planning stage can decide to directly answer the question.
    Question-blind subtasks replace the planned subtasks with three Subagents, each searching one third of the video for the question.
    Without Subagents, \method answers from the glance frames directly.
    With retrieved frames, it also sees the 80 frames that best match the question in CLIP.
    In a single context, the Root agent solves the subtasks itself instead of Subagents, and each of its turns starts with the glance or with the notes and kept frames.
    In the answering stage, we drop the reports, notes, reason, initial answer, or kept frames, or show the whole glance in place of the notes and kept frames.
    These rows rerun only the answering stage on the recorded planning stage and reports.
    Accuracy thus comes from the Subagent search, while planning, the notes, and separate contexts mainly save tokens.}
    \label{tab:ablation}
    \vspace{-0.05in}
    \footnotesize
    \setlength{\tabcolsep}{3pt}
    \begin{tabular*}{\linewidth}{@{\extracolsep{\fill}}lcc}
    \toprule
    Configuration & Acc.\,$\uparrow$ & Input $\downarrow$ \\
    \midrule
    \method & 59.7 & 46.0K \\
    \multicolumn{3}{l}{\emph{Planning stage}} \\
    \quad w/ early answer & 59.5 & 43.3K \\
    \quad w/ question-blind subtasks & 59.1 & 64.3K \\
    \multicolumn{3}{l}{\emph{Subagents}} \\
    \quad w/o Subagents & 52.4 & 30.8K \\
    \quad w/o Subagents, w/ retrieved frames & 55.0 & 46.2K \\
    \quad w/ single context, glance per turn & 58.8 & 163.5K \\
    \quad w/ single context, notes per turn & 58.4 & 51.2K \\
    \multicolumn{3}{l}{\emph{Answering stage}} \\
    \quad w/o reports & 53.7 & 32.9K \\
    \quad w/o notes & 58.9 & 45.9K \\
    \quad w/o reason & 59.6 & 45.9K \\
    \quad w/o initial answer & 59.4 & 46.0K \\
    \quad w/o kept frames & 59.3 & 45.1K \\
    \quad w/ whole glance & 60.2 & 72.3K \\
    \bottomrule
    \end{tabular*}
    \vspace{-0.1in}
\end{table}

We analyze how each part of \method affects accuracy and input tokens, with \backbone on LVBench (\cref{tab:ablation}).
The Subagent search contributes the most to accuracy, and the planned subtasks and the notes the most to efficiency.
Each ablation changes one part, and most reuse the recorded planning stage, so only that part differs.

\noindent \textbf{Root Planning.}
Early answering allows the agent to answer early without spanning subtasks. 
With an early answer, the planning stage ends the run on the 15\% of questions where the glance suffices.
It scores 59.5 and saves 2.7K input tokens per question. 
We hence conclude that whether to enable this setting or not does not significantly affect our method's performance and efficiency. 
Question-blind subtasks replace the planned subtasks with mutual 3 Subagents, each searching 1/3 of the video for the question.
They score 59.1 with 64.3K input tokens.
We conclude that the Root agent is better at self-dividing Subagents. 

\noindent \textbf{Subagent Search.}
Removing the Subagents decreases accuracy the most.
Without them, \method answers from the glance alone, and accuracy drops from 59.7 to 52.4.
Without Subagents, adding the frames that best match the question in embedding similarity raises the accuracy only to 55.0. 
It sees 80 such frames at 512 pixels, with the same input tokens as \method.
Unlike this one-shot retrieval, a Subagent searches with its own text, reads frames densely in the ranges it picks, and searches again when they do not settle the subtask.
The improvement thus comes from the targeted Subagent search (embedding search).

\noindent \textbf{Separate Contexts.}
Running each subtask in a separate context saves around 1/10 of the input tokens, with similar accuracy.
In \method, each Subagent starts a new context with only the question and its subtask.
In the single-context control, the Root agent solves the same subtasks itself with the same actions and frame budget.
Each turn starts with the glance or with the notes and kept frames, and the results of the earlier subtasks.
With the glance, it scores 58.8 with 3.6$\times$ the input tokens.
With the notes and kept frames, it scores 58.4 with 1.11$\times$ the input tokens.
Both controls use a single context.
Hence, we conclude that the glance frames make the first setting expensive.

\noindent \textbf{Answering Stage.}
The improvement over the initial answer is mainly from the Subagent reports.
We keep all the previous conversation and run only the answering stage again. 
Without the reports, it scores 53.7, close to the 52.4 without Subagents.
The reports thus increase it by 6.0 points.
Dropping the notes, reason, initial answer, or kept frames lowers the accuracy by at most 0.8 points.
It also saves at most 0.9K input tokens per question.
With the whole glance in place of the notes and kept frames, it scores 60.2 with 26K more input per question.
\method would then take 72.3K input tokens, more than the 65.6K total tokens of AVP.
Overall, the accuracy of \method mainly comes from the Subagent search, while planning, separate contexts, and the notes mainly save input tokens per question.

\begin{table}[t]
    \centering
    \caption{\textbf{Cost by Stage.}
    We report the average number across all questions in all three datasets. 
    Subtasks: subtasks dispatched.
    2--3: percentage of questions with two or three subtasks.
    Frames: frames that the Subagents read.
    Plan., Sub., Ans.: input tokens of the planning stage, the Subagents, and the answering stage.
    GPT-6-Sol and Qwen3.8-27B dispatch more subtasks than \backbone.}
    \label{tab:cost}
    \vspace{-0.05in}
    \footnotesize
    \setlength{\tabcolsep}{2pt}
    \begin{tabular*}{\linewidth}{@{\extracolsep{\fill}}lcccccc}
    \toprule
    & \multicolumn{3}{c}{Subagents} & \multicolumn{3}{c}{Input tokens} \\
    \cmidrule(lr){2-4} \cmidrule(lr){5-7}
    Benchmark & Subtasks & 2--3 & Frames & Plan. & Sub. & Ans. \\
    \midrule
    \multicolumn{7}{l}{\emph{Qwen3.8-27B}} \\
    \quad LVBench & 1.40 & 38.7 & 25.8 & 22.1K & 15.2K & 2.9K \\
    \quad Video-MME long & 1.69 & 59.3 & 33.4 & 22.0K & 19.8K & 3.4K \\
    \quad LVB long & 1.32 & 29.6 & 23.9 & 21.9K & 14.7K & 3.0K \\
    \multicolumn{7}{l}{\emph{\backbone}} \\
    \quad LVBench & 1.14 & 18.3 & 19.7 & 30.8K & 12.2K & 2.9K \\
    \quad Video-MME long & 1.52 & 49.8 & 29.6 & 30.5K & 17.7K & 3.6K \\
    \quad LVB long & 1.18 & 18.1 & 18.7 & 30.1K & 12.8K & 3.1K \\
    \multicolumn{7}{l}{\emph{GPT-6-Sol}} \\
    \quad LVBench & 1.76 & 58.0 & 32.0 & 30.3K & 17.9K & 3.5K \\
    \quad Video-MME long & 2.24 & 85.8 & 44.8 & 30.2K & 24.3K & 4.4K \\
    \quad LVB long & 1.62 & 49.6 & 26.5 & 30.1K & 15.5K & 3.4K \\
    \bottomrule
    \end{tabular*}
    \vspace{-0.1in}
\end{table}

\noindent \textbf{Subtasks.}
The planning stage consumes the most part of the input tokens, and GPT-6-Sol and Qwen3.8-27B dispatch more subtasks than \backbone (\cref{tab:cost}).
With \backbone, the Root agent dispatches 1.14 subtasks per LVBench question, and two or three for 18\% of the questions.
On Video-MME long, it dispatches two or three for half of the questions.
GPT-6-Sol dispatches two or three on 58\% of the LVBench questions, and its Subagents read 32.0 frames per question instead of 19.7.
Qwen3.8-27B dispatches 1.32 to 1.69 subtasks per question, between the two.
In its first round of conversation, each Subagent decides how it searches and reads.
For 1{,}368 of the 1{,}769 LVBench subtasks, the Subagent searches first.
It locates an object or person (1{,}069), finds when an event happens (150), finds all instances (127), or orders events (22).
For the rest, it reads a part of the video uniformly.

\noindent \textbf{Cost by Stage.}
The planning stage costs about the same on every benchmark, and the Subagents cost it differently (\cref{tab:cost}).
With \backbone, the planning stage costs 30.1K to 30.8K input tokens per question, as every glance has 256 frames.
The Subagents take 12.2K on LVBench and 17.7K on Video-MME long, where they run more subtasks.
The answering stage costs only 2.9K to 3.6K.
GPT-6-Sol costs 5.7K more Subagent input than \backbone on LVBench, and its Subagents add 10.6 points against 7.3.
Qwen3.8-27B counts fewer tokens per glance frame, and its planning stage costs only about 22K input tokens.
Its Subagents still cost more input than those of \backbone.

\noindent \textbf{Glance Coverage.}
On Video-MME long, the initial answer alone reaches 64.6, similar to AVP (\cref{tab:qwen}).
Its questions are mostly global, such as a synopsis, the order of events, or a count over the whole video.
A glance frame every 7 to \SI{14}{s} already shows the evidence for most of these questions.

\section{Conclusion}
\label{sec:conclusion}
\method plans from a low-resolution glance, lets Subagents search and read in new contexts, and answers from its notes and their reports.
It never shows the glance twice.
With \backbone, it matches AVP and VideoLucy on LVBench and Video-MME long with about 30\% and 90\% fewer tokens.
It also outperforms other methods on LongVideoBench long. 
Compared to uniform sampling baselines, with similar input tokens, it achieves 4.7 points higher accuracy. 
Based on different question types, different mechanisms contribute differently.  
For example, the Subagents contribute 7.3 to 10.6 \% accuracy on LVBench.
On Video-MME long, the glance alone contributes to the accuracy. 
Showing the glance frames only in the planning stage makes \method more efficient.
The Subagents' searching makes it accurate.
With stronger backbones such as GPT-6-Sol, \method widens its lead over AVP on LVBench from 1.1 to 5.0 points.

\noindent \textbf{Limitations.}
\method dispatches subtasks even when the glance settles the answer.
On Video-MME long, they change accuracy by at most 1.8 points with any backbone and cost about 21K input tokens per question.
Only \method and AVP run on other backbones, and no run uses reasoning.
The CLIP index costs GPU time when a video gets a single question.
The planning stage takes the largest share of the input with every backbone.
Future work can learn when to dispatch, compress the glance, retrieve events instead of frames, and train the Root agent to plan better subtasks from the low-resolution glance alone.

{\small
\bibliographystyle{ieeenat_fullname}
\bibliography{main}
}

\clearpage
\onecolumn
\appendix
\crefalias{section}{appendix}
\crefalias{subsection}{appendix}
\begin{center}
    {\Large\textbf{\thetitle}\par}
    \vspace{0.5em}
    {\Large Supplementary Material\par}
\end{center}
\vspace{1.0em}
{\large\bf
\begin{itemize}[itemsep=0.5em]
    \item \cref{sec:supp_alg}: Table of Notations
    \item \cref{sec:supp_details}: Method (\cref{sec:method})
    \item \cref{sec:supp_qa}: Long-Video Question Answering (\cref{sec:exp_setup})
    \item \cref{sec:supp_more}: Ablation Studies (\cref{sec:exp_ablation})
    \item \cref{sec:supp_agents}: Published Video Agents (\cref{sec:related,sec:exp_setup})
    \item \cref{sec:supp_prompts}: Prompts (\cref{sec:method})
    \item \cref{sec:supp_examples}: Qualitative Examples (\cref{fig:example})
\end{itemize}
}
\clearpage
\makeatletter\setlength{\@fptop}{0pt}\makeatother
\raggedbottom

\section{Table of Notations}
\label{sec:supp_alg}

{\small
\begin{longtable}{@{}llr@{}}
    \caption{\textbf{Table of Notations.}}
    \label{tab:notation}\\
    \toprule
    Notation & Description & Default \\
    \midrule
    \endfirsthead
    \multicolumn{3}{@{}l}{\cref{tab:notation}, continued.} \\
    \toprule
    Notation & Description & Default \\
    \midrule
    \endhead
    \bottomrule
    \endfoot
    $V$                                     & Input video & \\*
    $T$                                     & Duration of $V$ in seconds & \\*
    $t$                                     & Time in seconds, $t\in[0,T]$ & \\*
    $x_t$                                   & Frame of $V$ nearest to time $t$ & \\*
    $S$                                     & Set of distinct frames that the VLM sees for one question & \\*
    $q$                                     & Question & \\*
    $\mathcal{Y}$                           & Set of answer options & \\*
    $\hat{y}^\mathrm{init}$                 & Initial answer, given in the planning stage & \\*
    $\hat{y}^\mathrm{final}$                & Final answer, given in the answering stage & \\
    \midrule
    $\pi_\theta$                            & Frozen VLM & \\*
    $\theta$                                & Parameters of $\pi_\theta$ & \\*
    $\pi_\theta^{\mathrm{root}}$            & Root agent: $\pi_\theta$ with the prompts of the Root agent & \\*
    $\pi_\theta^{\mathrm{sub}}$             & Subagent: $\pi_\theta$ with the prompts of the Subagent & \\
    \midrule
    $\Gamma$                                & Glance: $G$ low-resolution frames at the middles of $G$ equal parts of $V$ & \\*
    $m$                                     & Notes of the planning stage on what happens in $V$ and when & \\*
    $\mathcal{K}$                           & Keep list of the planning stage: glance frames that the answering stage sees again & \\*
    $\mathcal{Q}$                           & Set of subtasks of the planning stage & \\*
    $u$                                     & Subtask: an instruction for one Subagent & \\*
    $u_k$                                   & The $k$-th subtask in $\mathcal{Q}$ & \\
    \midrule
    \textsc{Sub}$(u,q,\mathcal{Y})$         & Subagent procedure of \cref{alg:sub} for subtask $u$ & \\*
    $\ell$                                  & Turn of a Subagent, $\ell=1,\dots,R$ & \\*
    $o_\ell$                                & Observation of turn $\ell$; $o_1=(q,\mathcal{Y},u)$ & \\*
    $\tilde{o}_\ell$                        & Text version of $o_\ell$: its frame labels without the images & \\*
    $z_\ell$                                & Reply of turn $\ell$, $z_\ell=(m_\ell,a_\ell,\mathcal{K}_\ell)$ & \\*
    $m_\ell$                                & Reason of reply $z_\ell$ & \\*
    $a_\ell$                                & Action of reply $z_\ell$ & \\*
    $\mathcal{K}_\ell$                      & Keep list of reply $z_\ell$: earlier frames that turn $\ell+1$ sees again & \\*
    $\{(\tilde{o}_i,z_i)\}_{i<\ell}$        & Text versions and replies of the turns before turn $\ell$ & \\*
    $b$                                     & Remaining frame budget of a Subagent, starting at $B$ & \\
    \midrule
    \textsc{Search}$(c,w)$                  & Action that shows the strongest CLIP peaks for query $c$ in $w$ & \\*
    \textsc{Read}$(w,f)$                    & Action that shows frames with uniform spacing in $w$ at frame rate $f$ & \\*
    \textsc{Return}$(y,\mathcal{H},E,U)$    & Action that ends the Subagent and gives its report & \\*
    $w$                                     & Time window of a search or read, picked by the Subagent & \\*
    $t_0$                                   & Start time of $w$ & \\*
    $t_1$                                   & End time of $w$ & \\*
    $c$                                     & Text query of a search & \\*
    $k'$                                    & Maximum number of peaks that a search shows, $k'=\min(k,b)$ & \\*
    $f$                                     & Frame rate of a read, in frames per second & \\*
    $n$                                     & Number of frames that a read decodes (\cref{eq:read}) & \\*
    $E$                                     & Indices of the evidence frames that a return cites & \\
    \midrule
    $r$                                     & Report of a Subagent, $r=(y,\mathcal{H},\mathcal{E},U,\sigma)$ & \\*
    $r_k$                                   & Report for subtask $u_k$ & \\*
    $y$                                     & Free-text answer of the report & \\*
    $\mathcal{H}$                           & Times that $y$ refers to & \\*
    $\mathcal{E}$                           & Evidence frames: the frames of $E$, decoded again at their timestamps & \\*
    $U$                                     & Points that the frames of the Subagent do not settle & \\*
    $\sigma$                                & Flag set when the limits of the Subagent stopped it & \\
    \midrule
    $i$                                     & Index of a frame in the CLIP index & \\*
    $t_i$                                   & Time of indexed frame $i$, $t_i=i/\nu$ & \\*
    $\phi_{\mathrm{img}}$                   & CLIP image encoder & \\*
    $\phi_{\mathrm{txt}}$                   & CLIP text encoder & \\*
    $\mathbf{f}_i$                          & Normalized image feature of $x_{t_i}$ & \\*
    $s_i(c)$                                & Cosine similarity of indexed frame $i$ to query $c$ & \\
    \midrule
    $G$                                     & Number of glance frames, each the size of 64 tiles of $32\times32$ pixels & 256 \\*
    $M$                                     & Maximum number of subtasks per question & 3 \\*
    $n_g$                                   & Maximum number of glance frames in $\mathcal{K}$ & 16 \\*
    $B$                                     & Frame budget of a Subagent & 24 \\*
    $R$                                     & Maximum number of replies of a Subagent & 8 \\*
    $k$                                     & Maximum number of peaks per search & 4 \\*
    $\delta$                                & Minimum distance of two peaks & \SI{10}{s} \\*
    $n_r$                                   & Maximum number of frames per \textsc{Read} & 8 \\*
    $n_e$                                   & Maximum number of evidence frames per report & 3 \\*
    $n_k$                                   & Maximum number of frames in $\mathcal{K}_\ell$ & 4 \\*
    $\nu$                                   & Frame rate of the CLIP index (ViT-B/32) & \SI{1}{fps} \\
    \midrule
    $N$                                     & Number of frames of the Uniform-$N$ baseline, at the middles of $N$ equal parts of $V$ & \\
\end{longtable}
}

\FloatBarrier
\section{Method}
\label{sec:supp_details}
This section adds the rules that the code applies to the method of \cref{sec:method}, in the order of its subsections.

\subsection{Root Agent}
\label{sec:supp_root}
\noindent \textbf{Planning.}
Besides the outputs of \cref{eq:call1}, the reply of the planning stage holds a reason, which states what the glance settles and what the answer still needs.
The answering stage receives this reason with the notes and the initial answer.
A subtask is valid if its instruction is not empty.
\method keeps the first $M$ valid subtasks of the reply and drops the other entries.
It maps each number in the keep list $\mathcal{K}$ to the nearest glance frame, drops repeats, and keeps at most $n_g$ frames, in time order.

\subsection{Subagent}
\label{sec:supp_sub}
\noindent \textbf{Video Environment.}
CLIP has an image encoder $\phi_{\mathrm{img}}$ and a text encoder $\phi_{\mathrm{txt}}$.
For each video, we sample frames at times $t_i=i/\nu$ with $0\le t_i<T$ and store the normalized features $\mathbf{f}_i=\phi_{\mathrm{img}}(x_{t_i})/\lVert\phi_{\mathrm{img}}(x_{t_i})\rVert_2$.
A text query $c$, called a caption in \cref{sec:sub}, gives each indexed frame the score
\begin{equation}
s_i(c) \;=\; \mathbf{f}_i^{\top}\,\frac{\phi_{\mathrm{txt}}(c)}{\lVert\phi_{\mathrm{txt}}(c)\rVert_2}.
\end{equation}
\textsc{Search}$(c,w)$ shows at most $k'=\min(k,b)$ frames for query $c$ inside $w$, where $b$ is the remaining budget.
It averages the scores over a centered window of \SI{5}{s} and takes as peaks the frames whose smoothed score is the highest within $\delta/2$ seconds.
It scans the peaks in decreasing order of score and skips a peak closer than $\delta$ to a peak taken before.
A peak within $\delta/2$ of a frame that an earlier search of the same Subagent showed is taken but not shown.
It still suppresses its neighbors, and a repeated search shows the next candidates.
The scan stops at $k'$ shown peaks, which are listed in temporal order.
If no peak is left, the observation says so and the search costs nothing.
A read of $w=[t_0,t_1]$ at frame rate $f$ decodes
\begin{equation}
n \;=\; \max\bigl(1,\,\min(n_r,\,b,\,\lfloor(t_1-t_0)f\rfloor+1)\bigr)
\label{eq:read}
\end{equation}
frames with uniform spacing and costs $n$ units of the budget.
A capped read states its actual frame times and density.
A search costs one unit per frame shown.
An invalid reply costs a reply but no budget, and the next observation is an error message.

\noindent \textbf{Frames and Labels.}
A glance frame keeps its aspect ratio and covers the area of 64 tiles of $32\times 32$ pixels, the low spatial rate of the AVP code.
A 16:9 frame thus has a longer side of 341 pixels.
A local frame has a 512-pixel longer side.
All frames are JPEG images.
On \backbone, a glance frame costs about 115 input tokens with its label, and a local frame about 210.
A glance frame has the label \texttt{[glance | t=<seconds>s]}.
A Subagent frame has the label \texttt{[frame <index> | t=<seconds>s | <source>]}, where the source is \texttt{search} or \texttt{read}.
A kept frame gets the extra tag \texttt{kept}.
In the answering stage, an evidence frame has the label \texttt{[task <k> evidence | t=<seconds>s]}.
Model inputs contain no file paths or benchmark identifiers.

\noindent \textbf{Reports and Evidence Frames.}
\method checks each index in $E$ against the frames shown to the Subagent, removes repeats, keeps the first $n_e$ valid entries, and decodes their source frames into $\mathcal{E}$.
When the budget or the replies run out, the last reply must be a return.
Otherwise, the report takes the answer and evidence of that reply, sets $\sigma$, and lists the subtask itself among the unconfirmed points.

\subsection{Frame Budget and Reason-as-Memory}
\label{sec:supp_budget}
\noindent \textbf{Reason-as-Memory.}
The first turn of a Subagent holds no image, and kept frames cost no budget.

\noindent \textbf{Answering from Notes.}
The planning stage holds the $G$ glance frames.
The answering stage holds at most $n_g$ kept glance frames and $n_e$ evidence frames per subtask, at most $n_g+Mn_e=25$ images.
On the 1{,}442 LVBench questions with subtasks, it takes 3.1K input tokens, 1.0K of them for the kept glance frames.
With the whole glance instead of the notes and kept frames, it takes 31.4K.

\FloatBarrier
\section{Long-Video Question Answering}
\label{sec:supp_qa}
This section adds the details of \cref{sec:exp_setup}, under the names of its paragraphs.

\subsection{Benchmarks and Metrics}
\label{sec:supp_bench}
\noindent \textbf{Video Decoding.}
The decoder fetches frames by timestamp.
For time $t$, it takes the frame index nearest to $t$ times the average frame rate of the video.
We convert 71 LVBench videos to H.264 for decoder compatibility: 69 AV1 videos and 2 VP9 videos with seek failures.
The conversion uses CRF 18, keeps the source timestamps, and removes audio.
Each converted video is within \SI{0.17}{s} of the source duration.
The 300 Video-MME long videos are used as released, in H.264.

\noindent \textbf{MLVU and LongVideoBench.}
Besides the three benchmarks of \cref{tab:main}, we run \method on the 2{,}174 multiple-choice questions of the MLVU dev set~\cite{zhou2025mlvu}.
Its metric M-Avg is the mean accuracy over the question types.
In the dev set, 15 questions repeat a candidate, 5 of them the correct one; like the official scorer, we take the last matching candidate as the correct option.
We also run it on the other 773 questions of the LongVideoBench validation set, on videos of up to 10 minutes.

\subsection{Settings}
\label{sec:supp_impl}
\noindent \textbf{CLIP Index.}
We precompute one CLIP index per video with OpenCLIP~\cite{ilharco2021openclip}, using ViT-B/32 with OpenAI weights, and share it across all questions about that video.
Image features are computed on GPUs at \SI{1}{fps}, and a CPU encodes the text queries during question answering.

\noindent \textbf{Backbone Access.}
All methods run \backbone with reasoning effort ``none'', without tools, and with no prompt other than their own.
The system prompt of a method becomes the instructions of the request, and its user content the user turn, with the images in order.
All methods use the default sampling temperature and output limit of the API.
Every Root run and every Subagent turn is a new conversation.
The parser takes the first message of the reply that contains a JSON object.

\noindent \textbf{Failed Runs.}
A run is sent again when its reply holds no JSON object, is refused, fails, or stalls for \SI{240}{s}.
It is sent again at most twice in the runs before October 7 and at most five times after.
From October 20 on, a run that the API rate-limits also waits 30 to \SI{600}{s}, at most eight times, before it is sent again.
Questions that still fail on service errors run again, with the same rule for every method.
The tokens and model runs in \cref{tab:main,tab:uniform,tab:qwen,tab:ablation,tab:cost} count every attempt.

\noindent \textbf{Qwen3.8-27B.}
For \cref{tab:qwen}, vLLM 0.29.0 serves Qwen3.8-27B on one GPU, with thinking mode off as the counterpart of reasoning effort ``none''.
A request holds at most 512 images of up to 704 pixels, and the context has 128K tokens.
The model receives only the prompts of each method and replies in the JSON mode of vLLM.
It samples with its own generation configuration and writes at most 4{,}096 output tokens per reply.
A reply cut at this limit fails its question: 3 questions of \method on each benchmark and 10 LVBench questions of AVP.
\method keeps the prompts and settings of its \backbone runs and answers 16 questions at a time on one H100.
Its three benchmarks take 6.7 GPU hours, including the start-up of the server.

\noindent \textbf{GPT-6-Sol.}
For \cref{tab:qwen}, GPT-6-Sol runs \method and AVP with the access, the settings, and the prompts of the \backbone runs, also with reasoning effort ``none''.
One LVBench question of \method fails in every attempt, on a Subagent reply that ends in the middle of a sentence.

\subsection{Baselines}
\label{sec:supp_baselines}
\Cref{sec:exp_setup} runs every baseline with its own prompts and default settings, adapted where needed.
The paragraphs below give these adaptations.
A question counts as wrong when its final reply holds no valid answer.

\noindent \textbf{AVP.}
We keep the official decision procedure of AVP~\cite{wang2025active} and run it with \backbone.
The prompts, answer extraction, confidence threshold of 0.7, and three-round plan--observe--reflect loop are unchanged.
At the low spatial rate, a frame takes 64 visual tokens and an observation holds at most 512 frames.
At the medium rate, a frame takes 256 tokens and an observation holds at most 128 frames.
These caps are the defaults of its code, a quarter of those of its Gemini configuration (2{,}048 and 512 frames).
On LVBench, AVP stops after one round on 81\% of the questions and sees 417 frames per question, mostly at low resolution.

\noindent \textbf{LensWalk.}
We run the official code of LensWalk~\cite{li2026lenswalk} with its prompts and its standard tool configuration, at most 20 turns, with \backbone as every model.
We run \backbone without native tool calls.
The tool schemas and a JSON reply format are appended to the system prompt, and each model run receives the conversation so far as a transcript.
On 62, 34, and 28 questions of the three benchmarks, its final reply holds no valid option letter.
One further LVBench question fails in every attempt.

\noindent \textbf{VideoAtlas.}
We run the official code of VideoAtlas~\cite{eltahir2026videoatlas} with its prompts and default settings, with \backbone as the Master and every Worker.
The defaults are $8\times8$ grids of 336-pixel cells, a Master and seven Workers, and a budget of 512 model runs with 8 steps per cell.
Zooms go down to spans of one second.
As for LensWalk, the tool schemas and a JSON reply format go into the system prompt.
The official code requires the reply to use a tool whenever it offers tools, so the reply format asks for one.
Each grid of 2{,}688 pixels reaches \backbone downscaled to 1{,}600 pixels.
Its frames in \cref{tab:main} are the distinct frames placed in its grids.
Its Workers collect no evidence on 80, 39, and 32 questions of the three benchmarks, and on 110, 90, and 35 questions it answers that the evidence cannot decide.

\noindent \textbf{ReViSe.}
We run the plug-and-play loop of the official code of ReViSe~\cite{xu2026towards} with its prompts and settings, as in the driver of its LVBench and Video-MME experiments.
A question takes at most four rounds, each with at most three frames out of 20 candidates, at 384 pixels on the longer side.
An invalid reply is retried at most twice.
Every round covers the whole video: we do not pass the LVBench reference window, which its LVBench loader uses for prompts and frame sampling.
ReViSe does not cover LongVideoBench, and we use the same settings there.
Its replies are plain text with tags, and the reply is the last non-empty message of a run, because the model often writes a plain message first.

\noindent \textbf{VideoLucy.}
We run the question loop of the official code of VideoLucy~\cite{zuo2026videolucy} with its prompts and defaults and with \backbone as both captioner and LLM.
The coarse memory uses clips at \SI{1}{fps}: 100 to \SI{200}{s} long on LVBench, by the duration rule of its paper, and \SI{100}{s} long on Video-MME long.
LongVideoBench long, which its paper does not cover, uses the LVBench rule.
Fine and ultra-fine clips last 10 and \SI{1}{s} at \SI{2}{fps}, with at most five iterations.
The sparse initialization asks for two time periods as in the code; the paper says three.
A clip reaches \backbone as the frames that the Qwen2.5-VL reader of the code would take, at the same sizes, each labeled with its time in the clip.
On 43, 11, and 18 questions of the three benchmarks, the code stops on a reply that it cannot process, such as an empty time period.
One further LVBench question is stopped by the output filter of the API in every attempt.

\noindent \textbf{LLoVi.}
The official code of LLoVi~\cite{zhang2024simple} holds only the LLM stage, with its prompts for a 500-word summary and an answer from it.
For the captions, we follow the reproduction by DrVideo~\cite{ma2025drvideo}: one frame every \SI{5}{s}, with the prompt ``Describe the image.'', at 672 pixels on the longer side.
One model run shows 8 consecutive frames and asks for a separate description of each.

\noindent \textbf{VideoAgent.}
The official code of VideoAgent~\cite{wang2024videoagent} runs only on EgoSchema, from cached model replies, captions, and EVA-CLIP features.
We reproduce its question loop for any number of options with its prompts: five uniform frames first and at most three rounds of answering, self-evaluation, and retrieval.
Retrieval uses the CLIP index of \method (ViT-B/32, \SI{1}{fps}) instead of EVA-CLIP-8B-plus.
\backbone captions a frame when a step reads it, with the frames, size, and prompt of the LLoVi baseline, at most 8 frames per run.
We remove the lines of the prompts that define the marks of the EgoSchema captions.
The code draws a random answer when it cannot read a reply, which decides 8 of the 3{,}013 final answers; we seed the draw per question.

\noindent \textbf{Uniform Sampling and Blind.}
\Cref{sec:supp_prompts} gives the prompts of Uniform-$N$ and Blind.
Blind receives no duration and no frame.

\subsection{Results}
\label{sec:supp_results}
\noindent \textbf{LVBench by Capability and Video Category.}
\Cref{tab:breakdown} splits the LVBench results of \cref{tab:main} by the six capabilities of the benchmark and by video category.
\method is above AVP on entity recognition, event understanding, and key information retrieval, and below it on temporal grounding, reasoning, and summarization, none of them significantly.
Over its initial answer, \method improves the most on temporal grounding (+17.3 points) and key information retrieval (+14.1).
It does not improve on reasoning or summarization.
Uniform-400 scores only 46.8 on temporal grounding, where the question names a time.

\begin{table}[!htp]
    \centering
    \caption{\textbf{LVBench Accuracy (\%) by Capability and Video Category} with \backbone, all 1{,}549 questions.
    ER: entity recognition. EU: event understanding. KIR: key information retrieval. TG: temporal grounding. Rea: reasoning. Sum: summarization.
    A question can have several capabilities.
    Corrected / lost: questions that the answering stage turns from wrong to right / from right to wrong; $^*$initial answer.}
    \label{tab:breakdown}
    \vspace{-0.05in}
    \begin{tabular}{lcccccc}
    \toprule
    Capability & ER & EU & KIR & TG & Rea & Sum \\
    (questions) & (677) & (647) & (291) & (220) & (201) & (58) \\
    \midrule
    Uniform-400 & 56.7 & 53.5 & 61.2 & 46.8 & 55.2 & 43.1 \\
    AVP & 58.3 & 56.6 & 61.5 & 70.0 & 57.7 & 53.4 \\
    \method{}$^*$ & 53.0 & 51.5 & 51.9 & 50.5 & 48.8 & 55.2 \\
    \method & 60.6 & 58.3 & 66.0 & 67.7 & 51.2 & 51.7 \\
    Corrected / lost & 85 / 34 & 80 / 36 & 50 / 9 & 47 / 9 & 20 / 15 & 6 / 8 \\
    \midrule
    Category & Self-m. & Live & TV & Cartoon & Sport & Docum. \\
    (questions) & (317) & (262) & (250) & (283) & (240) & (197) \\
    \midrule
    Uniform-400 & 56.8 & 53.8 & 54.0 & 53.4 & 55.8 & 56.3 \\
    AVP & 61.5 & 61.1 & 54.8 & 56.2 & 55.8 & 61.9 \\
    \method{}$^*$ & 50.2 & 54.2 & 53.6 & 49.8 & 50.4 & 57.9 \\
    \method & 59.0 & 60.7 & 60.0 & 60.1 & 56.2 & 62.4 \\
    \bottomrule
    \end{tabular}
\end{table}

\noindent \textbf{MLVU and LongVideoBench.}
With the settings of \cref{tab:main}, \method reaches an M-Avg of 74.6 on the MLVU dev set.
On the other 773 LongVideoBench validation questions, it scores 76.1, which gives 73.1 on all 1{,}337 validation questions.
\Cref{tab:agents} compares both scores with published agents.

\subsection{Uniform Sampling}
\label{sec:supp_uniform}
\Cref{tab:uniform_other} extends \cref{tab:uniform} to three more benchmarks.
At similar input tokens, \method is significantly above Uniform-400 on LongVideoBench long and on MLVU dev.
On these two benchmarks, Uniform-256 is 5.7 and 4.7 points lower than \method, respectively.
Uniform-450 reads 12\% and 18\% more input tokens than \method, and it is 2.7 and 5.1 points lower.
On Video-MME long, \method and AVP are both similar to uniform sampling with 256, 400, or 450 frames.
With 222 uniform frames at 512 pixels, at the input tokens of \method, uniform sampling scores 55.0 on LVBench, as Uniform-400 does.

\begin{table}[!htp]
    \centering
    \caption{\textbf{Comparison with Uniform Sampling on Three More Benchmarks} with \backbone, all questions.
    Acc: accuracy in \%, M-Avg on MLVU dev.
    Input: input tokens per question.}
    \label{tab:uniform_other}
    \vspace{-0.05in}
    \begin{tabular}{lcccccc}
    \toprule
    & \multicolumn{2}{c}{Video-MME long} & \multicolumn{2}{c}{LongVideoBench long} & \multicolumn{2}{c}{MLVU dev} \\
    \cmidrule(lr){2-3} \cmidrule(lr){4-5} \cmidrule(lr){6-7}
    Method & Acc.\,$\uparrow$ & Input $\downarrow$ & Acc.\,$\uparrow$ & Input $\downarrow$ & Acc.\,$\uparrow$ & Input $\downarrow$ \\
    \midrule
    Uniform-256 & 62.8 & 29.5K & 63.3 & 29.5K & 69.9 & 30.8K \\
    Uniform-400 & 64.4 & 46.1K & 64.4 & 46.0K & 70.7 & 48.0K \\
    Uniform-450 & 63.3 & 51.8K & 66.3 & 51.7K & 69.5 & 53.9K \\
    \method & 64.3 & 51.8K & 69.0 & 46.0K & 74.6 & 45.6K \\
    \bottomrule
    \end{tabular}
\end{table}

\subsection{Backbones Analysis}
\label{sec:supp_backbones}
With Qwen3.8-27B in \cref{tab:qwen}, \method takes 7.9, 9.5, and 7.6 model runs per question on LVBench, Video-MME long, and LongVideoBench long and shows 280, 287, and 275 distinct frames.
With GPT-6-Sol, it takes 9.0, 11.2, and 8.1 model runs and shows 286, 298, and 278 distinct frames.

\FloatBarrier
\section{Ablation Studies}
\label{sec:supp_more}
This section defines the settings of \cref{tab:ablation} and adds results, under the paragraph names of \cref{sec:exp_ablation}.
A replay keeps the recorded planning stage of the default run, and its costs include that stage.

\subsection{Root Planning}
\label{sec:supp_planning}
\noindent \textbf{Early Answer.}
With an early answer, the planning stage also decides whether the glance alone settles the answer.
If the glance frames clearly show the evidence that singles out one option, the reply says so.
Its initial answer is then final: no subtask runs and there is no answering stage.
Otherwise, the planning stage gives subtasks as in the default run.

\noindent \textbf{Question-Blind Subtasks.}
This control replaces the planned subtasks with three equal parts of the video, each with the question as its instruction and the budget $B=24$.
It does so also on the questions without a planned subtask.
It replays the planning stage, so the notes, the initial answer, and the kept frames stay fixed.
\Cref{tab:segments} gives its accuracy and costs on the three benchmarks.
It is similar to \method in accuracy on each of them.
It takes more model runs, input tokens, and Subagent frames, as the Root agent dispatches only 1.14, 1.52, and 1.18 subtasks per question.

\begin{table}[!htp]
    \centering
    \caption{\textbf{Question-Blind Subtasks on Three Benchmarks} with \backbone, all questions.
    Acc: accuracy in \%.
    Runs: model runs per question.
    Input: input tokens per question.
    Frames: frames that the Subagents read per question.}
    \label{tab:segments}
    \vspace{-0.05in}
    \setlength{\tabcolsep}{3.5pt}
    \begin{tabular}{lcccccccccccc}
    \toprule
    & \multicolumn{4}{c}{LVBench} & \multicolumn{4}{c}{Video-MME long} & \multicolumn{4}{c}{LongVideoBench long} \\
    \cmidrule(lr){2-5} \cmidrule(lr){6-9} \cmidrule(lr){10-13}
    Configuration & Acc. & Runs & Input & Frames & Acc. & Runs & Input & Frames & Acc. & Runs & Input & Frames \\
    \midrule
    \method & 59.7 & 6.7 & 46.0K & 19.7 & 64.3 & 8.5 & 51.8K & 29.6 & 69.0 & 7.0 & 46.0K & 18.7 \\
    \quad w/ question-blind subtasks & 59.1 & 14.6 & 64.3K & 47.2 & 63.4 & 14.3 & 65.0K & 50.7 & 67.0 & 16.7 & 71.5K & 56.1 \\
    \bottomrule
    \end{tabular}
\end{table}

\subsection{Subagent Search}
\label{sec:supp_search}
\noindent \textbf{Retrieved Frames.}
Without Subagents, \method gives the initial answer of the default run.
With retrieved frames, the 80 frames are the peaks of the Subagent search (\cref{sec:supp_sub}) over the whole video, with the question text as the query.
One model run sees the glance and then these frames, each in time order.

\subsection{Separate Contexts}
\label{sec:supp_contexts}
\noindent \textbf{Single Context.}
In both single-context controls, the Root agent carries out the subtasks of the recorded planning stage itself, one after another.
With the glance per turn, every turn begins with the glance, the question, and the reports of the earlier subtasks.
With notes per turn, the glance is replaced by what the answering stage receives: the question, the reply of the planning stage, and the kept glance frames.
On the 283 LVBench questions with two or three subtasks, the control with notes takes 70.8K input tokens against 60.2K for \method.
With GPT-6-Sol on 500 random LVBench questions, it scores 72.6 against 70.2 for \method, with 58.4K against 52.0K input tokens.

\subsection{Answering Stage}
\label{sec:supp_answering}
\noindent \textbf{Replays on LVBench.}
The answering-stage settings of \cref{tab:ablation} are replays.
The reply of the planning stage and the reports stay fixed, and only the answering stage runs again, under the same system prompt.
A second run with every input scores 59.5 and gives the answer of the default run on 92.6\% of the 1{,}442 questions with subtasks.
Without the notes, the reason, the initial answer, or the kept glance frames, the answering stage stays within 0.6 points of this second run.
It gives the answer of that run on 86.5 to 92.9\% of these questions.
Only the reports change the accuracy significantly.

\noindent \textbf{Other Benchmarks.}
We repeat the replays on LongVideoBench long and Video-MME long, where 15 and 97 questions have no valid subtask and keep the initial answer.
A second run with every input scores 68.8 and 64.4, against 69.0 and 64.3 for \method.
Without the reports, the answering stage scores 68.4 and 64.7.
The reports thus change the accuracy by $+0.6$ and $-0.4$ points, against $+6.0$ on LVBench.
On Video-MME long, the two runs with every input agree on 91.2\% of the 803 questions with subtasks.

\subsection{Subtasks}
\label{sec:supp_subtasks}
On LVBench, the Subagents return on their own in all 1{,}769 subtasks, and none is stopped by its budget or its replies.

\subsection{Cost by Stage}
\label{sec:supp_cost}
\Cref{tab:cost} leaves out the CLIP index, which is built once per video.
The index of the 205.5 hours of Video-MME long videos took 52.6K seconds of A100 jobs with 8 CPU cores, or 256\,s per video hour.
Decoding, at 14 frames per second, bounds this time.
Shared by the questions of a video, the index costs about 19, 58, and 34\,s of GPU time per question on LVBench, Video-MME long, and LongVideoBench long.

\subsection{Glance Coverage}
\label{sec:supp_glance}
\noindent \textbf{Video-MME Long by Question Type.}
On the 900 Video-MME long questions, \method is above AVP on temporal reasoning (65.9 against 57.1, 91 questions) and object reasoning (65.8 against 64.2, 240 questions).
It is below AVP on action reasoning (53.9 against 60.6, 180 questions), information synopsis (80.4 against 84.0, 163 questions), and OCR (50.0 against 78.6, 14 questions).
Its answering stage is below the initial answer on object reasoning (65.8 against 67.9) and counting (47.9 against 52.1).

\FloatBarrier
\section{Published Video Agents}
\label{sec:supp_agents}
\Cref{sec:exp_setup} compares \method with seven training-free video agents with public code, and \cref{sec:related} groups video agents by their input.
This section explains how we found these agents and assigned their input types.
A method counts as a video agent if a model gathers evidence from the video over several steps, or if a language model answers from captions of the video.
We found 96 video agents with a score on LVBench or on the long split of Video-MME in a standard setting.
The score comes from their own paper or, for two of them, from another paper.
\Cref{tab:agents} lists the 21 of them that are training-free, have public code, and report a score without subtitles, speech, or audio.

\begin{table}[!htp]
    \centering
    \caption{\textbf{Accuracy (\%) of Training-Free Video Agents with Public Code on Four Long-Video Benchmarks.}
    LVBench has 1{,}549 questions.
    The Video-MME column is its long split of 900 questions, the MLVU column is its dev set, scored by the mean accuracy over its multiple-choice question types (M-Avg), and the LongVideoBench column is its validation set.
    The input of an agent is F if a VLM inspects frames that the agent selects for the question, and C if a language model reasons over captions or a text memory of the video.
    The caption model writes the captions or the text memory, and the agent model plans, reasons, and answers.
    Each agent uses its strongest single-run setting with named models that uses no subtitles, speech, or audio.
    $^{*}$ marks LVBench scores on a subset of the 1{,}549 questions, $^{\dagger}$ marks scores that another paper reports for the agent, and -- marks models that are not reported and scores that are not reported or that use subtitles, speech, or audio.
    Bold marks the best unmarked score in each column.}
    \label{tab:agents}
    \vspace{-0.05in}
    \footnotesize
    \setlength{\tabcolsep}{3pt}
    \begin{tabularx}{\linewidth}{@{}lc>{\raggedright\arraybackslash}p{2.9cm}>{\raggedright\arraybackslash}Xcccc@{}}
    \toprule
    Method & Input & Caption Model & Agent Model & LVBench & Video-MME & MLVU & LongVideoBench \\
    \midrule
    AVP~\cite{wang2025active} & F & & Gemini-2.5-Pro & \textbf{74.8} & \textbf{81.9} & \textbf{84.3} & \textbf{73.4} \\
    LensWalk~\cite{li2026lenswalk} & F & & o3 & 68.6 & 71.4 & -- & -- \\
    VideoDeepResearch~\cite{yuan2025videodeepresearch} & F & & DeepSeek-R1 + Seed1.5-VL-Pro & 55.5 & -- & -- & -- \\
    T$^3$~\cite{huang2026temporal} & F & & LLaVA-Video-7B & 47.3 & 54.2 & -- & 61.4 \\
    VideoAtlas~\cite{eltahir2026videoatlas} & F & & Gemini-3-Flash & -- & 76.2 & -- & -- \\
    LVAgent~\cite{chen2025lvagent} & F & & InternVL2.5-78B/8B + Qwen2-VL-72B & -- & 74.3 & 83.9 & -- \\
    A4VL~\cite{xu2026multi} & F & & InternVL3-78B +\newline LLaVA-Video/Qwen2.5-VL-72B & -- & 68.3 & -- & -- \\
    VideoChat-A1~\cite{wang2026videochat} & F & & InternVideo2.5-8B & -- & 65.0 & 76.2 & 65.4 \\
    \midrule
    DVD~\cite{zhang2026deep} & F+C & GPT-4.1/4.1-mini & o3 & 74.2 & 67.3 & -- & 71.6 \\
    MERIT~\cite{choi2026keep} & F+C & GPT-5-mini & GPT-5 & 71.8 & -- & -- & -- \\
    VideoSeek~\cite{lin2026videoseek} & F+C & GPT-5 & GPT-5 & 68.4 & 70.1 & -- & -- \\
    PACE~\cite{xu2026finding} & F+C & Qwen3-VL-30B-A3B-Thinking & Qwen3-VL-30B-A3B-Thinking & 62.6 & 58.9 & -- & -- \\
    AVA~\cite{yan2026ava} & F+C & Qwen2.5-VL-7B & Qwen2.5-32B + Gemini-1.5-Pro & 62.3 & 64.1 & -- & -- \\
    MR.~Video~\cite{pang2026mr} & F+C & Gemini-2.0-Flash & GPT-4o + Gemini-2.0-Flash & 60.8$^{*}$ & 63.4 & -- & -- \\
    VideoLucy~\cite{zuo2026videolucy} & F+C & Qwen2.5-VL-7B & DeepSeek-R1 & 58.8$^{*}$ & 66.8 & 76.1 & -- \\
    DrVideo~\cite{ma2025drvideo} & F+C & LLaVA-NeXT-7B & DeepSeek-V2.5 & -- & 51.7 & -- & -- \\
    \midrule
    VideoAgent~\cite{wang2024videoagent} & C & Qwen2-VL & GPT-4o & 29.3$^{\dagger}$ & 46.4$^{\dagger}$ & -- & -- \\
    SiLVR~\cite{zhang2025silvr} & C & NVILA-7B & DeepSeek-R1 & -- & 62.7 & -- & -- \\
    VideoHV-Agent~\cite{wang2026thinkverify} & C & -- & GPT-4o & -- & 60.6 & -- & -- \\
    VideoTree~\cite{wang2025videotree} & C & LLaVA-NeXT-7B & GPT-4 & -- & 54.2 & 60.4 & -- \\
    LLoVi~\cite{zhang2024simple} & C & LLaVA-NeXT-7B & DeepSeek-V2.5 & -- & 45.4$^{\dagger}$ & -- & -- \\
    \midrule
    \method (ours) & F & & \backbone & 59.7 & 64.3 & 74.6 & 73.1 \\
    \bottomrule
    \end{tabularx}
\end{table}

\noindent \textbf{Input Types.}
We label each agent by the inputs that its paper describes, with F and C as in the caption of \cref{tab:agents}.
For F, the VLM is the answering model or a tool of the agent.
A VLM that only captions frames with a fixed prompt counts toward C.
In the counts below, caption-based means C or F+C, and frame-based means F.
\method is of type F: its Root agent sees a uniform glance, its Subagents read the frames that they select for their subtasks, and no step captions the video.
Most caption-based agents also inspect frames for the question.
DVD and PACE use a frame-inspection tool, AVA checks the frames of the retrieved events before it answers, and MERIT gives the answering model the frames of retrieved clips.
MR.~Video, VideoLucy, and DrVideo prompt a VLM with the question or an instruction derived from it on selected scenes, clips, or frames.
VideoSeek also has its model describe uniformly sampled frames of the whole video without the question.
VideoAgent, SiLVR, VideoTree, and LLoVi answer from captions alone, and the code of VideoHV-Agent captions frames with a fixed prompt.

\noindent \textbf{Selection.}
An agent has public code if its paper or project page links a repository that contains its implementation.
We checked the frame-based agents on 30 September 2026 and the caption-based agents on 1 October 2026.
A training-free agent does not fine-tune its VLMs or LLMs.
LVAgent and VideoChat-A1 fine-tune only the CLIP model that they use for retrieval, so we count them as training-free.
We leave out 41 agents without public code (27 frame-based, 14 caption-based) and 26 agents with public code that fine-tune a VLM or LLM (20 frame-based, 6 caption-based).
We also leave out 8 training-free agents with public code whose scores on the four benchmarks all use subtitles, speech, or audio.
The code release lists all 96 agents with the reason for each one left out.

\noindent \textbf{Scores.}
The papers of VideoAgent and LLoVi report none of the four benchmarks.
VCA~\cite{yang2025vca} reruns VideoAgent with GPT-4o and Qwen2-VL captions, DrVideo reruns LLoVi with DeepSeek-V2.5 and LLaVA-NeXT-7B captions, and \cref{tab:agents} gives these scores.
VideoLucy and MR.~Video evaluate 1{,}492 LVBench questions on 99 of the 103 videos.
MERIT uses speech transcripts on Video-MME, VideoDeepResearch uses subtitles on Video-MME and LongVideoBench, and LVAgent and A4VL use them on LongVideoBench, so \cref{tab:agents} leaves out these scores.
DVD captions LVBench with GPT-4.1 and the other benchmarks with GPT-4.1-mini, VideoTree uses LLaVA-OV-7B captions on MLVU, and VideoHV-Agent does not name its captioner for Video-MME.
AVA states its scores only in the text, and we give its default models.
The scores of MR.~Video are those of its NeurIPS version.

\noindent \textbf{MLVU and LongVideoBench Splits.}
MLVU has a dev set with four options per question and a test set with six, and \cref{tab:agents} lists only dev-set scores.
We identify the split of each score by the evaluation code or the number of questions where possible.
Otherwise, we use the GPT-4o score in the same table, which is 64.6 on the dev set and 54.5 on the test set~\cite{zhou2025mlvu}.
AVP names its set the test split, but the set has the multiple-choice questions of the dev set (2{,}175 by its count; the released files have 2{,}174).
A4VL and VideoDeepResearch report only the test set, and VideoTree names the dev set the validation set.
The LongVideoBench column is the validation set of 1{,}337 questions, and \cref{sec:supp_results} gives the score of \method on it.
LensWalk, VideoSeek, VideoAtlas, MR.~Video, and PACE report only its long subset of videos of 900 to 3{,}600 seconds, so their cells are empty.
We leave out a LongVideoBench score when the paper states that subtitles or transcripts are part of the input.

\FloatBarrier
\section{Prompts}
\label{sec:supp_prompts}
The boxes below reproduce the prompts of \method with the default parameters, and the prompts of the Uniform-$N$ and Blind baselines of \cref{sec:exp_setup}.
The Root prompt holds both stages of \cref{sec:root_plan}, and the Subagent prompt the actions of \cref{sec:sub} and the memory of \cref{sec:budget}.
Fields in braces are filled in at run time and are set in purple.
The prompts call a subtask a task.
They call the planning and answering stages call 1 and call 2, and the text that a search gives to CLIP a query.
The system prompt of the Root agent also calls a Subagent a worker.

\csname nolinenumbers\endcsname
\begin{promptbox}{root}{Root}{$\pi_\theta^{\mathrm{root}}$}
\promptpart{System prompt}{both stages}
\begin{lstlisting}[style=prompt]
You are the Root of a video question-answering system. You answer a multiple-choice question about a long video in two calls.

You see a glance of the whole video: frames sampled uniformly over its full length, at low resolution, in time order, each labelled `[glance | t=<seconds>s]`. The glance shows what the video is about and how it is structured in time, but it is sparse: what happens between two glance frames is not shown, and small details may not be readable. You and every task see video frames only, with no audio and no subtitles: a task can only ask about what is visible (people, actions, objects, scenes, on-screen text).

Call 1, plan. The glance is shown only in this call: in call 2 you get your notes and at most 16 glance frames that you choose, not the whole glance. Write notes on the glance and give your initial answer: the option that the glance supports best, even if you are unsure. Then give 1 to 3 local tasks. A task is an instruction about the video. A worker that has not seen the glance carries out the task by itself, anywhere in the video. It searches the video with a CLIP image-text search (frames that look like a short caption), reads parts of the video at a higher resolution than the glance, densely around a candidate moment or spread over a part, under a budget of 24 frames and at most 8 steps, and then reports its answer, the times it refers to, and its evidence frames. What each task found comes back to you in call 2.
{"notes": "<the video in time order: its parts with their time ranges, and what each shows that bears on the question and the options>", "reason": "<what the glance settles, and what the answer still needs>", "answer": "<option letter>", "keep": [<t of a glance frame>, ...], "tasks": [{"task": "..."}, ...]}
- notes: your memory of the glance in call 2. Write down what the answer may need: people, objects, scenes, on-screen text, counts and the order of events, each with its time.
- keep: the times of at most 16 glance frames that matter most for the answer; you see them again in call 2.
- task: a self-contained instruction saying what to find and what to report (for example the order of steps, a count with the time of each instance, the text on a sign, whether the event of an option happens). Name the candidates from the options that must be told apart. Whenever the task names a person, object, scene or on-screen text, describe what it looks like in the glance (clothing, colours, setting, layout): the worker never sees the glance and searches by appearance, so a target named only by its role ("the host", "the winner") cannot be found.

Call 2, answer: you see your call-1 notes and initial answer, the glance frames you kept, your tasks, and what each task found. Answer the question from all of it. Your initial answer came from sparse, low-resolution frames, and the tasks looked closer: where a task result shows evidence for or against an option, it outweighs the glance; where the task results do not bear on the question, your initial answer stands.
{"reason": "<brief reasoning over your notes and the task results>", "answer": "<option letter>"}

Every reply is one JSON object and nothing else.
\end{lstlisting}
\tcblower
\begin{promptmsg}
\promptpart{User message}{planning stage}
\begin{lstlisting}[style=prompt]
Video duration: {duration}s
Glance: {glance} frames, one every {spacing}s:
<256 glance frames, each after its label [glance | t=<seconds>s]>
Question: {question}
Options:
{options}
This is call 1: give your notes, your initial answer, the glance frames to keep and 1 to 3 tasks. Reply with one JSON object.
\end{lstlisting}
\end{promptmsg}
\medskip
\begin{promptmsg}
\promptpart{User message}{answering stage}
\begin{lstlisting}[style=prompt]
Video duration: {duration}s
The glance of call 1 had {glance} frames, one every {spacing}s; it is not shown again.
Question: {question}
Options:
{options}
Your call-1 reply (notes on the glance, initial answer): {reply}
The glance frames you kept:
<kept glance frames, each after its label [glance | t=<seconds>s]>
This is call 2: answer the question. Your tasks from call 1 and what they found:
Each task went to a Subagent: a separate model that saw only the question, the options and its task, not the glance, and that searched and read the video by itself. Its report gives the strategy it chose, its answer to the task, the times it refers to, what it could not confirm, how many frames it saw, and whether it was stopped by its limits before it returned by itself, followed by at most 3 evidence frames that it selected, labelled `[task <k> evidence | t=<seconds>s]`. A report is a claim from one Subagent: check it against its evidence frames and your notes. The reasoning across the glance and all tasks is yours.
Task {k}: {task}
Report: {report}
<evidence frames, each after its label [task <k> evidence | t=<seconds>s]>
Reply with one JSON object: your answer.
\end{lstlisting}
\promptnote{The subtask block repeats for every subtask. The reply field holds the notes, reason, and initial answer of the planning stage; with the kept glance frames, it replaces the glance.}
\end{promptmsg}
\end{promptbox}

\begin{promptbox}{sub}{Subagent}{$\pi_\theta^{\mathrm{sub}}$}
\promptpart{System prompt}{}
\begin{lstlisting}[style=prompt]
You are a Subagent of a video question-answering system. Another agent, which has seen a sparse glance of the whole video, gave you one task about the video. You also see the multiple-choice question and its options, as context, so that you know which details matter; your job is the task, not the question.

You look at the video only through actions. Every frame you are shown has a label `[frame <index> | t=<seconds>s | <source>]`; the index counts all frames shown to you, starting at 1, and the source is `search` or `read`. Every frame you are shown costs 1 budget. Every action stays inside the video: a time range you give is clipped to the video, and null means the whole video.

Each reply is ONE JSON object and nothing else: a "reason" (what the new frames show, with their frame indices, then what is known and what is missing) plus exactly one action. Your first reply also has "strategy": one of "locate", "when", "order", "find_all", "cover" (see below). Every frame is shown to you only once: at each later reply you get the earlier turns as text (the observations with their frame labels, and your replies) without their images, so "reason" is your only memory of what a frame showed. To see frames again, add "keep": [<frame index>, ...] to a search or read reply: at most 4 frames you have seen, from any turn; each list replaces the previous one. Kept frames cost no budget.

1. search: frames that look like a short caption.
{"reason": "...", "action": "search", "query": "...", "t0": <sec or null>, "t1": <sec or null>}
- A CLIP image-text retriever compares query with every second of the video (of the time range you give) and shows the 4 strongest distinct candidate moments, at least 10 s apart, in time order. It leaves out every moment within 5 s of a frame that an earlier search showed you, so searching again, even with the same query, shows the next candidates.
- query: a short caption of what the wanted frames LOOK like (visible people, objects, scene, on-screen text). CLIP matches appearance, so do not write questions, negations, or temporal words.

2. read: frames spread uniformly over a time range.
{"reason": "...", "action": "read", "t0": <sec>, "t1": <sec>, "fps": <number>}
- fps: frames per second, sampled uniformly inside [t0, t1]. One read shows at most 8 frames and never more than your remaining budget; if (t1 - t0) x fps + 1 is larger, the frames are spread over [t0, t1] at a lower density instead. The observation states the actual frame times. Narrow [t0, t1] to get the requested fps.

3. return: your report.
{"reason": "...", "action": "return", "answer": "<answer to the task>", "times": [<sec>, ...], "evidence": [<frame index>, ...], "unconfirmed": ["<what the task asks that your frames do not settle>"]}
- times: the times, in seconds, of the moments your answer refers to, e.g. each instance you count or each step you order ([] if none).
- evidence: the indices of at most 3 frames that best show your answer. They are handed to the agent that gave you the task.
- unconfirmed: each part of the task that your frames do not settle ([] if none).

Choose the strategy by the task:
- locate (who is this, what model or object, does X appear): search with a caption of its appearance, then read around the best candidate to confirm it.
- when (when does X happen): search for X, then read densely around the candidate to pin the moment down.
- order (does A happen before B): search for A and search for B, then read the local process around each.
- find_all (how many times, list every instance): do not stop at the first candidates. After each cluster of candidates, search again (the next search shows other moments), until a search shows nothing relevant; or read the video part by part. Give one time per distinct instance, and count an instance seen in several frames once.
- cover (what happens in this part): read uniformly over that part of the video; do not search.

Limits: a budget of 24 frames and at most 8 replies. When the budget is 0, or at your last reply, no further action is executed and the reply must be return.

Report only what you see in frames. Never guess: what your frames do not show goes in "unconfirmed".
\end{lstlisting}
\tcblower
\begin{promptmsg}
\promptpart{User message}{first turn}
\begin{lstlisting}[style=prompt]
Question the whole system is answering (context only):
{question}
Options:
{options}
Your task: {task}
Video duration: {duration}s
Frame budget: 24. At most 8 replies.
You have not seen any frame yet.
Frame budget left: 24. Replies left: 8. Reply with one JSON object.
\end{lstlisting}
\promptnote{Each later turn starts with ``Earlier turns (their frames are not shown again, except the frames you kept):'', the earlier observations and replies as text, the kept frames, and the new observation.}
\end{promptmsg}
\end{promptbox}

\begin{promptbox}{plain}{Uniform-sampling baseline}{}
\begin{promptmsg}
\promptpart{System prompt}{}
\begin{lstlisting}[style=prompt]
You answer a multiple-choice question about a video. You see frames sampled uniformly over the whole video, in time order. Every frame has a label `[frame <index> | t=<seconds>s]`, where t is the time of the frame in seconds from the start of the video.

Reply with one JSON object:
{"reason": "<brief reasoning>", "answer": "<option letter>"}
\end{lstlisting}
\end{promptmsg}
\tcblower
\begin{promptmsg}
\promptpart{User message}{}
\begin{lstlisting}[style=prompt]
Video duration: {duration}s
{n} frames sampled uniformly over the whole video:
<frame labels and frames>
Question: {question}
Options:
{options}
Reply with one JSON object.
\end{lstlisting}
\promptnote{The labeled frames are inserted at \prompttt{<frame labels and frames>}.}
\end{promptmsg}
\end{promptbox}

\begin{promptbox}{plain}{Blind baseline}{}
\begin{promptmsg}
\promptpart{System prompt}{}
\begin{lstlisting}[style=prompt]
You answer a multiple-choice question about a video that you cannot see: you get only the question and the options. Choose the option that is most likely correct, also when you are unsure.

Reply with one JSON object:
{"reason": "<brief reasoning>", "answer": "<option letter>"}
\end{lstlisting}
\end{promptmsg}
\tcblower
\begin{promptmsg}
\promptpart{User message}{}
\begin{lstlisting}[style=prompt]
Question: {question}
Options:
{options}
Reply with one JSON object.
\end{lstlisting}
\end{promptmsg}
\end{promptbox}
\csname linenumbers\endcsname

\FloatBarrier
\section{Qualitative Examples}
\label{sec:supp_examples}
\begin{figure}[!htp]
\centering
\includegraphics[width=\linewidth]{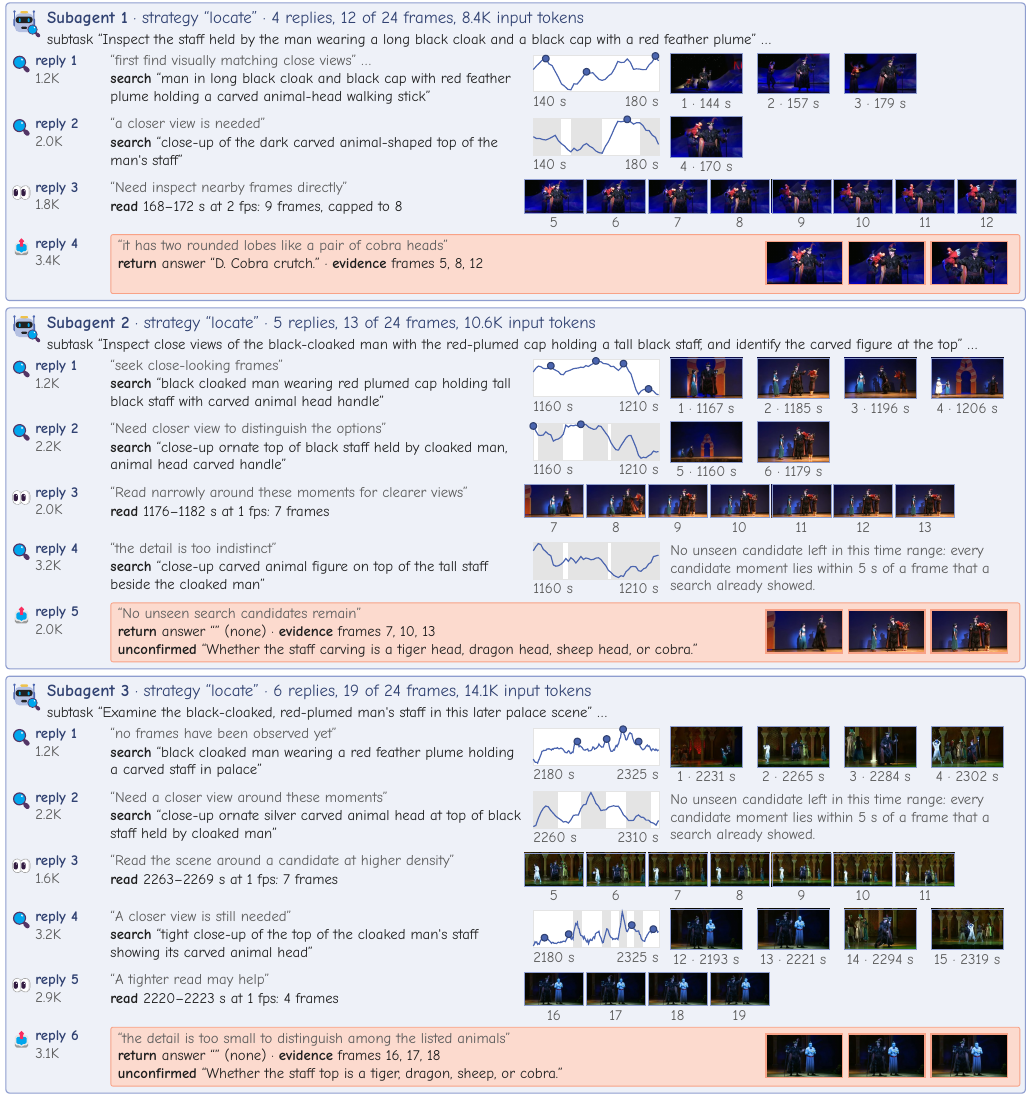}
\vspace{-0.2in}
\caption{\textbf{The Three Subagents of an LVBench Question, Reply by Reply.}
The frames are images that the model received, and the quotes are verbatim excerpts of the recorded subtasks and replies.
Each reply is a new model run that sees the earlier turns as text and the new frames, under a budget of 24 frames.
Each row gives the reply, its input tokens, an excerpt of its reason (gray), and its action.
For a search, the curve is the smoothed CLIP similarity of the query in the searched span, the dots are the peaks that become new frames, and the gray spans lie within \SI{5}{s} of frames that earlier searches showed, so the search leaves them out.
Frames are numbered in the order in which the Subagent sees them.
The orange box is the report, and code copies its evidence frames with unchanged pixels and timestamps.}
\label{fig:step_sub}
\end{figure}

\Cref{fig:example} shows a recorded run with one subtask, and \cref{fig:step_sub} a run with three, both of \method with \backbone on LVBench.
In \cref{fig:step_sub}, the question asks which crutch the man in the cloak holds during a stage show.
The glance shows an animal-headed staff but not which animal, and the Root agent gives the initial answer A, a tiger head.
It dispatches three subtasks that inspect the staff.
Subagent 1 reports a cobra head on the staff, and Subagents 2 and 3 return no answer.
The answering stage changes the answer to the correct D, a cobra crutch.

\end{document}